%% file: main.tex
\documentclass[sigconf, nonacm]{acmart}

\usepackage{enumitem}
\usepackage{float}
\usepackage{booktabs}
\usepackage{multirow}
\usepackage{graphicx}
\usepackage{xcolor}
\usepackage{subcaption}
\usepackage{lscape}
\usepackage{framed}
\usepackage{rotating} 
\usepackage{pifont}

\definecolor{shadecolor}{rgb}{0.9, 0.9, 0.9} 

\AtBeginDocument{%
  }

\copyrightyear{2026}
\acmYear{2026}
\setcopyright{cc}
\setcctype{by}
\acmConference[CHI '26]{Proceedings of the 2026 CHI Conference on Human Factors in Computing Systems}{April 13--17, 2026}{Barcelona, Spain}
\acmBooktitle{Proceedings of the 2026 CHI Conference on Human Factors in Computing Systems (CHI '26), April 13--17, 2026, Barcelona, Spain}
\acmPrice{}
\acmDOI{10.1145/3772318.3791835}
\acmISBN{979-8-4007-2278-3/2026/04}

\begin{document}

\title{Emulating Aggregate Human Choice Behavior and Biases with GPT Conversational Agents}

\author{Stephen Pilli}
\email{stephen.pilli@ucdconnect.ie}
\orcid{0000-0003-1655-1782}
\affiliation{%
  \institution{University College Dublin}
  \city{Dublin}
  \country{Ireland}
}

\author{Vivek Nallur}
\orcid{0000-0003-0447-4150}
\email{vivek.nallur@ucd.ie}
\affiliation{%
  \institution{University College Dublin}
  \city{Dublin}
  \country{Ireland}
}

\renewcommand{\shortauthors}{Pilli \& Nallur}

\begin{abstract}

Cognitive biases often shape human decisions. While large language models (LLMs) have been shown to reproduce well-known biases, a more critical question is whether LLMs can predict biases at the individual level and emulate the dynamics of biased human behavior when contextual factors, such as cognitive load, interact with these biases.
We adapted three well-established decision scenarios into a conversational setting and conducted a human experiment (N=1100).
Participants engaged with a chatbot that facilitates decision-making through simple or complex dialogues.
Results revealed robust biases.
To evaluate how LLMs emulate human decision-making under similar interactive conditions, we used participant demographics and dialogue transcripts to simulate these conditions with LLMs based on GPT-4 and GPT-5.
The LLMs reproduced human biases with precision. We found notable differences between models in how they aligned human behavior. This has important implications for designing and evaluating adaptive, bias-aware LLM-based AI systems in interactive contexts.

\end{abstract}



\maketitle

\section{Introduction}

Simulating human behavior has long been a goal of research in Human-Computer Interaction (HCI), cognitive science, and behavioral modeling ~\citep{hwang2025human}. 
Large language models (LLMs) have been fine-tuned to produce fluent, human-like dialogue and frequently achieve high performance on established benchmarks ~\citep{yi2024survey}.
This advancement presents opportunities beyond conventional applications by enabling simulation of large-scale human behavior in both experimental and policy contexts~\cite{park2023generative, park2024generative, argyle2023out, aher2023using, hamalainen2023evaluating}. 
To accurately represent human behavior in simulations, LLMs must demonstrate both human-like decision-making as well as the ability to reproduce human cognitive biases and irrationalities~\citep{binz2024turning, ying2025benchmarking}.
By empirically investigating the capacity of LLMs to emulate or accurately represent biased human behavior, they can be positioned as potential facilitators of realistic simulations. 

In its simplest form, decision-making requires a decision scenario and a set of alternatives to choose from. The structure and presentation of these alternatives can potentially shape decision-making by tapping into underlying cognitive biases. 
Cognitive biases are systematic deviations from rational judgment, arising from heuristics, prior experiences,
emotions, or social factors.~\citep{kahneman_thinking_2011}. Over 200 such biases have been systematically cataloged and experimentally validated through standardized cognitive tasks  ~\citep{big_bad_bias}. Notably, these tasks typically employ hypothetical scenarios that are detached from the complexities and nuances of real-world contexts.

Dialogue is a natural mode for facilitating decision-making using conversational agents. 
Within interactions with conversational agents, human dialogues often reveal cognitive biases, particularly in decision-making scenarios following cognitive tasks like structure~\citep{yamamoto_suggestive_2024, pilli_exploring_2024, ji_towards_2024}.
Among cognitive biases, Status quo bias is the most common and extensively explored. 
Status quo bias is the strong tendency for individuals to prefer current conditions over objectively better alternatives.
Rather than evaluating all options impartially, people anchor on the existing state and require disproportionately strong evidence or clear dominance before they are willing to switch. This bias can stem from loss aversion, cognitive effort avoidance, regret aversion, or simple inertia. This often leads decision-makers to stick with default choices or familiar options even when better alternatives exist~\citep{masatlioglu_rational_2005, samuelson_status_1988}.
Recent studies indicate that LLMs demonstrate well-documented cognitive biases when presented with cognitive tasks designed to elicit such effects~\cite{echterhoff2024cognitive, atreides2024cognitive, malberg2024comprehensive}.

Our goal diverges from the previous works; instead of investigating whether LLMs have biases, we examine whether LLMs can represent human behavior that is susceptible to cognitive biases. Such that, their decision-making while emulating an individual within a simulated environment - such as minimal, real-world task-oriented dialogue - reproduces biases. 

According to dual-process theory, decision-making involves two cognitive systems: \textit{System 1}, which is fast, intuitive, and heuristic-driven, and \textit{System 2}, which is slow, deliberate, and analytical~\citep{kahneman_thinking_2011}.
Cognitive biases can be understood as systematic deviations that arise from the interplay of these two systems, either from the over-reliance on System 1 heuristics or from failures or constraints in System 2 processing. 
Importantly, this interplay does not occur in isolation: contextual factors in the environment shape how strongly each system contributes to a decision. 
Features such as time pressure, task complexity, emotional state, or information presentation can shift the balance between System 1 and System 2. This, in turn, influences or interacts with the effect of the cognitive bias. 
One of the most common contextual factors that interacts with cognitive bias is cognitive load, which is the mental effort required for a task or dialogue~\citep{sweller1988cognitive, deck2015effect, brachten2020ability, litman2002designing, bogdanov2023working}.
Thus, cognitive load arising from the dialogue complexity of the task-oriented dialogue may interact with an individual's decision-making, potentially influencing individuals aggregate susceptibility to cognitive biases.

Given the context-dependent dynamics of cognitive biases, merely reproducing biased responses to cognitive tasks at the surface level is insufficient. Such reproduction does not adequately capture whether LLMs can faithfully represent human decision-making biases.
Therefore, we posit that LLMs should be able to reproduce biased behavior in a way that reflects sensitivity to contextual factors.
Their decision-making should not only reflect human biases but also align probabilistically with the biases observed in human decision-making while interacting with the context.
Despite a growing interest in LLM-based human behavioral modeling~\citep{binz2024turning, ying2025benchmarking}, no prior work was found in this line of inquiry.
We address this gap through a step-by-step empirical investigation, focusing on the question 
of \textit{How well do LLMs represent human biased decision-making behavior within a simulated conversational context, particularly when contextual factors are at play?}

To address this question, first, we establish a human baseline by examining whether humans exhibit the Status quo bias~\citep{samuelson_status_1988} when decision-making is facilitated through a conversational agent's dialogue. 
Second, we investigate the role of contextual factors within the dialogue. Specifically, the complexity of prior dialogue (the conversational turns that come before the decision-making) is a source of cognitive load that would interact with the decision-making biases. 
Third, we emulate human participants by creating LLM agents using demographic information and human-likeness prompts, following the method proposed by ~\citet{park2024generative}. We then compare each LLM agent's responses to the corresponding human responses to assess the predictive power of LLMs when leveraging individual-specific cues from the chat transcript.
This evaluation addresses a key question: Does the biased behavior represented by LLMs arise from individual-specific cues found in the prior dialogue?
Finally, to address our central research question. We examine the probabilistic alignment between biases observed in simulated and human responses at the group level. By comparing how biases manifest under varying conditions, such as dialogue complexity, we assess the degree to which LLMs not only reproduce cognitive biases but also capture their interaction with contextual factors.
Additionally, we analyze different human-likeness prompts to determine how much explicit instruction is needed for LLMs.

\section{Related Work}
\label{sec:Related Works}

Our work involves three key areas: cognitive bias in decision-making, LLM simulation, and conversational agents.

\subsection{LLM Simulation and Cognitive Biases}

Early work demonstrates that LLM-driven agents can exhibit coherent, human-like behavioral patterns and can be embedded in interactive simulations ~\citep{park2023generative, park2024generative}. Prior works also explored LLMs for social science simulation and large-scale societies ~\citep{hewitt2024predicting, piao2025agentsociety, beck2024sensitivity}. ~\citet{aher2023using} proposed a test to investigate how different language models can reproduce classic economic, psycholinguistic, and social psychology experiments: Ultimatum Game, Garden Path Sentences, Milgram Shock Experiment, and Wisdom of Crowds. In addition, ~\citet{binz2024turning} attempts to turn LLMs into cognitive models suggest that large, pre-trained models can be adapted to become models of human cognition.

Parallel lines of work examine whether LLM-based agents capture the cognitive limitations and systematic deviations from rationality that characterize human decision-making. 
Recent economic and behavioral studies argue that LLMs often assume overly rational agents compared to real humans ~\citep{liu2025rationality, kitadai2025can}. At the same time, large-scale evaluations show that LLMs can reproduce aggregate human cognitive bias patterns, including anchoring, framing, and mental accounting in single-shot settings~\citep{hwang2025human, malberg2024comprehensive, echterhoff2024cognitive, horton_large_2023, leng2024can}.
Systematic reviews further document the breadth and variability of such biases across models ~\citep{tjuatja2024llms}. 
These findings motivate recent efforts toward benchmarking human-likeness more broadly, including cognitive bias modeling, that is, the systematic capture of cognitive biases manifested in LLM simulations, as a crucial aspect for simulation ~\citep{ying2025benchmarking}

As this is an emerging field, only a few studies have explored LLM agents' capabilities to emulate irrationalities in a simulation. These works explored the reproduction of cognitive biases using only isolated prompts in single-shot or non-interactive settings. For instance, ~\citet{liu2025rationality} test LLM agents on basic cognitive tasks, such as choosing between risky options or inferring preferences, and use simple zero-shot and (chain-of-thought) prompts to examine how these models respond. Similarly, ~\citet{horton_large_2023}, which our work is closest to, evaluates LLM agents using straightforward economic decision tasks, such as dictator games, status-quo choices, fairness judgments, and hiring scenarios. Their method relies on clear, structured natural-language prompts of cognitive tasks to LLMs and elicits consistent responses from the models.
However, little is known about whether such cognitive biases emerge, persist, or interact with dialogue, which is an essential modality for human-like behavior given the established role of language in shaping high-level cognition ~\citep{hauser2002faculty}. This gap is especially important for HCI LLM simulation, where LLM agents are expected to show human-like decision biases during conversation rather than in isolated answers. 
Our work addresses this by examining how LLM agents reproduce cognitive biases within a conversational context and how well they emulate these biases while interacting with dialogue context. Such an investigation will overcome the limitation of superficial reproduction of biases in existing single-shot evaluations. Our investigation with LLMs required a reproducible and rigorous human baseline to compare. In order to identify a human baseline, we first review the existing works that investigate how cognitive biases manifest in human–conversational agent interactions.

\subsection{Cognitive Biases and Conversational Agents}

Cognitive biases in human decision-making have been examined closely in the fields of Psychology, Behavioral Economics, and Human-Computer Interaction. Recent research has examined their potential to both leverage and mitigate such biases~\citep{caraban_23_2019} using conversational agents. The following is a brief review of the works that potentially can be leveraged for our primary investigation.

~\citet{yamamoto_suggestive_2024} introduced ``suggestive endings'' in chatbot dialogue, based on the Ovsiankina effect. This design prompted users to engage more deeply, ask follow-up questions, and reflect longer, enhancing cognitive engagement.
~\citet{pilli_exploring_2024} used chatbots to assess cognitive biases like framing effects and loss aversion. Participants exhibited typical bias responses, confirming chatbots as valuable tools for bias detection and measurement.
~\citet{dubiel_impact_2024} examined the role of synthetic voice fidelity in decision-making. They found that high-fidelity voices, through cues like pitch and pace, enhanced source credibility and triggered affect heuristics, subtly influencing user choices.
Building on earlier work, 
~\citet{ji_towards_2024} studied cognitive biases in spoken conversational search (SCS), highlighting biases like anchoring and confirmation bias in the absence of visual cues. Their framework is largely theoretical but sets the stage for future bias-mitigation strategies in voice-based systems.
~\citet{ali_mehenni_nudges_2021} explored children's susceptibility to tasks resembling cognitive tasks used to infer cognitive biases by conversational agents and robots using a modified Dictator Game. Their findings revealed a stronger influence from artificial interlocutors than humans, pointing to authority and social influence biases, especially among vulnerable users. 
~\citet{kalashnikova_linguistic_nodate} investigated linguistic nudges promoting ecological behavior. By leveraging biases such as status quo bias and social conformity, they showed that chatbots and robots were more persuasive than humans in shaping opinions.

However, these studies come with certain limitations. \citet{pilli_exploring_2024}, achieve stronger experimental control but rely on preliminary or underdeveloped designs. Work by \citet{yamamoto_suggestive_2024, ali_mehenni_nudges_2021, kalashnikova_linguistic_nodate} focuses on less popular biases, limiting broader impact. \citet{dubiel_impact_2024} and \citet{ji_towards_2024} confined their works to single modality. Across all the studies reviewed, the importance of experiments designed with reproducibility and experimental control is emphasized, which are critical for advancing our investigation.
Together, these gaps necessitate a thorough empirical investigation into how cognitive biases manifest in human–conversational agent interactions.

\section{Hypotheses}
\label{sec:Hypothesis Development}

Our study proceeds in three stages, moving from human behavior to LLM prediction. Accordingly, we propose the following hypotheses:

\begin{enumerate}[label=\textbf{H\arabic*}, leftmargin=*]

    \item \textbf{(Human Baseline – Status Quo Bias).} 
    In a chatbot-mediated decision scenario, participants will be more likely to select an option framed as the \textit{status quo} than the same option framed as a \textit{non-status quo} alternative.

    \item \textbf{(Status Quo Bias - Complex Dialogue).} 
    Participants exposed to a cognitively demanding \textit{Complex Dialogue} prior to the decision scenario will demonstrate a stronger status quo bias than those who engage in a \textit{Simple Dialogue}.

    \item \textbf{(LLM Prediction – Reproducing Human Biases).} 
    When provided with demographic information and prior dialogue transcripts, LLMs will reproduce the behavioral patterns observed in H1 and H2. Specifically, they will predict both the baseline status quo bias and its variation under complex dialogue, across different levels of human-likeness prompting at the individual level.

\end{enumerate}

\section{Human Experiments}

To test the first two hypotheses, we design a dialogue that follows a real-world task-oriented dialogue structure and facilitates decision-making, yet ensures experimental control and ecological validity by standardizing dialogue content and controlling for confounding variables. 
A formal representation of our dialogue $\mathcal{D}$ with prior dialogue and decision scenario is as follows:

\begin{equation}
\nonumber
\mathcal{D} = 
\textbf{\{}
u_1^{sys}, u_2^{usr}, \ldots
\underbrace{ u_{t-k}^{sys}, \ldots , u_{t-1}^{usr},}_{\text{Prior Dialogue}}
\underbrace{u_t^{sys},u_{t+1}^{usr} }_{\substack{\text{Decision} \\ \text{Scenario} \\ \text{and} \\ \text{Response}}} \ldots
\textbf{\}}
\end{equation}

In this section, we start by outlining the decision scenarios adapted, explaining how we integrate them into the chatbot while maintaining consistency with prior research.
Next, we describe the Prior Dialogue tasks, which present either Simple or Complex Dialogues before the decision scenario. The details of the tasks are discussed in detail.
Further, we present the procedure and experiment design along with the questionnaire involved to measure aspects of the dialogue.

\subsection{Decision Scenarios}
\label{sec:Method-Decision Scenarios and Experimental Conditions}

The chatbot is designed to have introductory utterances like greetings 
$\{u_1^{sys},.., u_3^{sys},\ldots\}$$\in\mathcal{D}$
, which are followed by prior dialogue  
$\{u_{t-k}^{sys}, \ldots , u_{t-1}^{usr}\}$$\in\mathcal{D}$, 
and which is then followed by a decision scenario 
$ u_t^{sys}$$\in\mathcal{D}$
.
A decision scenario typically involves a decision-making problem accompanied by a set of alternatives from which participants must choose. In traditional experimental designs, a between-subjects approach is used to investigate the Status Quo bias. A control group is presented with a neutral version of the scenario where no alternative is designated as the status quo, while the treatment groups encounter scenarios where one of the options is explicitly framed as the status quo. The Status Quo effect is identified by analyzing the statistical difference in participant responses ${ u_t^{usr} } \in \mathcal{D}$ across these groups. Building on this methodology, we adapt a three-condition design for our experiments: a neutral condition, where neither option is framed as the status quo; a Status Quo A condition, where the first option is presented as the status quo; and a Status Quo B condition, where the second option is framed as the status quo. 

Our experiments adapted three decision scenarios from the original study by ~\citet{samuelson_status_1988}, namely: Budget Allocation (BA), Investment Portfolios (IDM), and College Job selection (CJ). 
These scenarios were chosen because they demonstrated a consistent Status Quo bias in the original experiments as well as in a recent replication study by ~\citet{xiao_revisiting_2021} in 2021. The change in the modality from pen and paper in ~\citet{samuelson_status_1988} to ~\citet{qualtrics2024} - a web-based tool to conduct online surveys - did not affect the replication. Moreover, they represent domains that are both widely studied in behavioral economics and highly relevant to practical applications in chatbot-based e-commerce and decision support systems. 

The following illustrates one of the decision scenarios and its corresponding experimental conditions.

\begin{framed}
\subsubsection*{\textbf{Chatbot Utterance $ u_t^{sys}$$\in\mathcal{D}$ for College Jobs Scenario (Neutral Condition)}}
$\\$
Having just completed your graduate degree, you have two offers of teaching jobs in hand.
When evaluating teaching job offers, people typically consider the salary, the reputation of the school, the location of the school, and the likelihood of getting tenure (tenure is a permanent job contract that can only be terminated for cause or under extraordinary circumstances).
Your choices are:
\begin{itemize}
 \item College A: East coast, very prestigious school, high salary, fair chance of tenure.
 \item College B: West coast, low prestige school, high salary, good chance of tenure.
\end{itemize}
\end{framed}

\begin{framed}
\subsubsection*{\textbf{Chatbot Utterance $ u_t^{sys}$$\in\mathcal{D}$ for College Jobs Scenario (Status Quo  A Condition - College A is status quo)}}
$\\$
You are currently an assistant professor at College A in the east coast. Recently, you have been approached by colleague at other university with job opportunity.
When evaluating teaching job offers, people typically consider the salary, the reputation of the school, the location of the school, and the likelihood of getting tenure (tenure is permanent job contract that can only be terminated for cause or under extraordinary circumstances).
Your choices are:
 \begin{itemize}
     \item Remain at College A: East coast, very prestigious school, high salary, fair chance of tenure.
     \item Move to College B: West coast, low prestige school, high salary, good chance of tenure.
 \end{itemize}
\end{framed}

\begin{framed}
\subsubsection*{\textbf{Chatbot Utterance $ u_t^{sys}$$\in\mathcal{D}$ for College Jobs Scenario (Status Quo B Condition  - College B is status quo)}}
$\\$
You are currently an assistant professor at College B in the west coast. Recently, you have been approached by colleague at other university with job opportunity.
When evaluating teaching job offers, people typically consider the salary, the reputation of the school, the location of the school, and the likelihood of getting tenure (tenure is permanent job contract that can only be terminated for cause or under extraordinary circumstances). 
Your choices are:
 \begin{itemize}
     \item Remain at College B: west coast, low prestige school, high salary, good chance of tenure.
     \item Move to College A: east coast, very prestigious school, high salary, fair chance of tenure.
 \end{itemize}
\end{framed}

In the Neutral Condition, neither College A nor College B are framed as a status quo alternative. Whereas, in Status Quo A condition, College A is framed as status quo, whereas in Status Quo B condition, College B is framed as status quo. The scenario for College B is the same, except that the status quo condition is reversed, \textit{i.e.,} the user is prompted that they are currently in College B and would like to move to College A.

Similar to the College Jobs (CJ) decision scenario, the Budget Allocation (BA) decision scenarios involve allocating the National Highway Safety Commission's budget between automobile safety and highway safety programs, with different conditions. 
The alternatives presented vary based on a neutral condition, where both allocation options are presented equally, and status quo conditions, where the current allocation either 60\% \textbf{A}uto safety / 40\% \textbf{H}ighway safety (60A40H) or 50\% \textbf{A}uto safety / 50\% \textbf{H}ighway safety (50A50H) influences whether respondents choose to maintain the existing budget or shift funds between programs. 

In the Investment Decision Making (IDM) scenario, choices are presented between a moderate-risk investment (Mod. Risk/ Company A) and a high-risk investment (High Risk/ Company B). Under neutral conditions, both options are presented without any prior commitments, while in status quo conditions, the investor inherits a portfolio already allocated to either moderate-risk or high-risk investments, potentially influencing their preference to maintain the existing investment or switch. 
The details of decision scenarios and their textual adjustments are available in Appendix ~\ref{sec:appendix-ds} and ~\ref{appendix:decision_scenario-manipulations}, respectively.

\subsection{Prior Dialogue}
\label{sec:Prior-dialogue}

A key feature of conversational choice environments is the presence of dialogue that precedes the decision scenario. We refer to this as the prior dialogue, denoted as $\{u_{t-k}^{sys}, \ldots , u_{t-1}^{usr}\} \in \mathcal{D}$, where $\mathcal{D}$ represents the full dialogue history. Our second hypothesis is that this part of the dialogue can potentially play a role in shaping subsequent decision-making. To investigate this, we designed two types of preference elicitation tasks: one that facilitates low-effort interaction, referred to as the \textit{Simple Dialogue}, and another that introduces cognitively demanding dialogue, referred to as the \textit{Complex Dialogue}. The dialogues were designed prioritizing experimental control while maintaining ecological validity. The characteristics of these tasks are as follows:

\subsubsection{Simple Dialogue}
A preference elicitation task was used for prior dialogue to facilitate a task-oriented dialogue. In the Simple Dialogue condition, participants are engaged in a set of short binary (yes/no) closed questions about preferences within a familiar domain. This task design was inspired by the schema-guided dialogue dataset (SGD) introduced by~\citet{rastogi_towards_2020} and aimed for a natural, low-effort interaction with the chatbot. Importantly, the Simple Dialogue used a conservative dialogue strategy: questions were direct, unambiguous, and did not require reasoning or memory beyond the current turn ~\citep{brachten2020ability, litman2002designing}. This ensured that the cognitive load imposed by the dialogue was minimal.

An example of the Simple Dialogue in the ``Music'' domain is shown below. A full list of domains and associated questions can be found in Section ~\ref{sec:appendix-simple_task-prior_dialogue} of Appendix ~\ref{sec:appendix-prior_dialogue}.

\begin{table}[htpb]
\centering
\caption[]{Simple Dialogue Attributes and Respective Utterances}
\label{tab:simple_dialogue_example}
\begin{tabular}{ll}
\toprule
\multicolumn{1}{c}{\textbf{Attribute}} & \multicolumn{1}{c}{\textbf{Yes/No Question}} \\ \midrule
Genre Preference & Do you like listening to pop music? \\
Language of Lyrics & \begin{tabular}[c]{@{}l@{}}Do you prefer music with lyrics in \\ English?\end{tabular} \\
Live Performances & \begin{tabular}[c]{@{}l@{}}Are you interested in live music \\ performances?\end{tabular} \\
Instruments Focused & Do you enjoy instrumental music? \\
Artist-Specific & \begin{tabular}[c]{@{}l@{}}Do you like music from specific artists? \\ Please enter ``I don't know'' only.\end{tabular} \\
Era (e.g., 80s, 90s) & Do you prefer music from the 90s? \\ 
\bottomrule
\end{tabular}%
\end{table}

Participants were instructed to answer each question directly and were prompted to enter ``I don't know'' for \textit{prior dialogue attention check} on one attribute. 
This low-complexity preference elicitation task served as a neutral prior dialogue before the decision scenario.

\subsubsection{Complex Dialogue}

The Complex Dialogue was designed to induce cognitive load in a controlled yet ecologically valid manner. To achieve this, a nested referential structure is used to manage multiple interdependent entities across a multi-turn dialogue.
The following shows an example of the utterances from $u^{sys}_{t-k}$ to $u^{sys}_{t-2}$ by the chatbot during the prior dialogue.
\begin{enumerate}
    \item[$u^{sys}_{t-8}$:] The first artist performs three live shows, is paid 2000 units per show, and has a 4-star rating. The second artist performs twice as many shows, with the same pay and rating. Which artist do you prefer, and why?
    \item[$u^{sys}_{t-6}$:] The third artist performs the same number of shows as the second, earns half the pay of the first artist, but has the same rating as the first. Which artist do you prefer, and why?
    \item[$u^{sys}_{t-4}$:] The fourth artist performs the same number of shows as the second, earns the same pay as the third, but has two stars less than the first artist. Which artist do you prefer, and why?
    \item [$u^{sys}_{t-2}$:] Remember the details of the fourth artist. Specific information will be requested later.
\end{enumerate}

The fourth artist in the chat transcript is defined by the second and third artists, which reference the second and first artists, creating a chain of dependencies. 
This design necessitates users to maintain and integrate hierarchical relationships between entities, thereby increasing semantic integration costs and working memory demands. 

This aligns with psycholinguistic findings that nested dependencies elevate processing difficulty
~\citep{gibson1998linguistic, van1983strategies}.
Additionally, the reappearance of earlier referents after intervening turns results in high referential distance~\citep{chen1985discourse}, which further taxes memory retrieval processes~\citep{arnold1998reference}.
From a cognitive load design standpoint, this structure directly aligns with Sweller's Cognitive Load Theory (CLT), which distinguishes between intrinsic load (task-related complexity), extraneous load (inefficient information presentation), and germane load (effort used for schema building)~\citep{sweller1988cognitive, deck2015effect}. 
From a dialogue systems perspective, managing multiple entities simultaneously while maintaining coherent context is a well-documented challenge, particularly when entities are interconnected or revisited~\citep{ultes2020complexity, asri2017frames}. 

Our Complex Dialogue incorporates these principles within a preference elicitation task.
As this approach requires making arithmetic comparisons between attributes and memorizing outcomes, it theoretically demands greater mental effort from the individual.
Similarly structured dialogues are used in other domains, including artist recommendations, streaming services, calendar apps, and banking options.
For the remaining complex dialogues, please refer to Table ~\ref{tab:appendix-complex-task} in Section ~\ref{sec:appendix-complex_task-prior_dialogue} of Appendix ~\ref{sec:appendix-prior_dialogue}.

\subsection{Technical Details of the Chatbot}

The web-based experimental interface was developed using Streamlit~\cite{streamlit}, which allowed for easy deployment and consistent access across devices. 
GPT-4o Mini, a large language model ~\citep{openai2024chatgpt4o}, was used to create realistic and coherent chatbot interactions based on structured prompts designed for each condition. An example of the prompt used for the agent is provided in Appendix~\ref{sec:appendix-prompt}.
The code for chatbot used in (SQB) experiment is available at
\url{https://github.com/stephen-pilli/exp-status-quo-bias.git}.

\begin{figure*}[ht]
    \centering
    \includegraphics[width=1.0\linewidth]{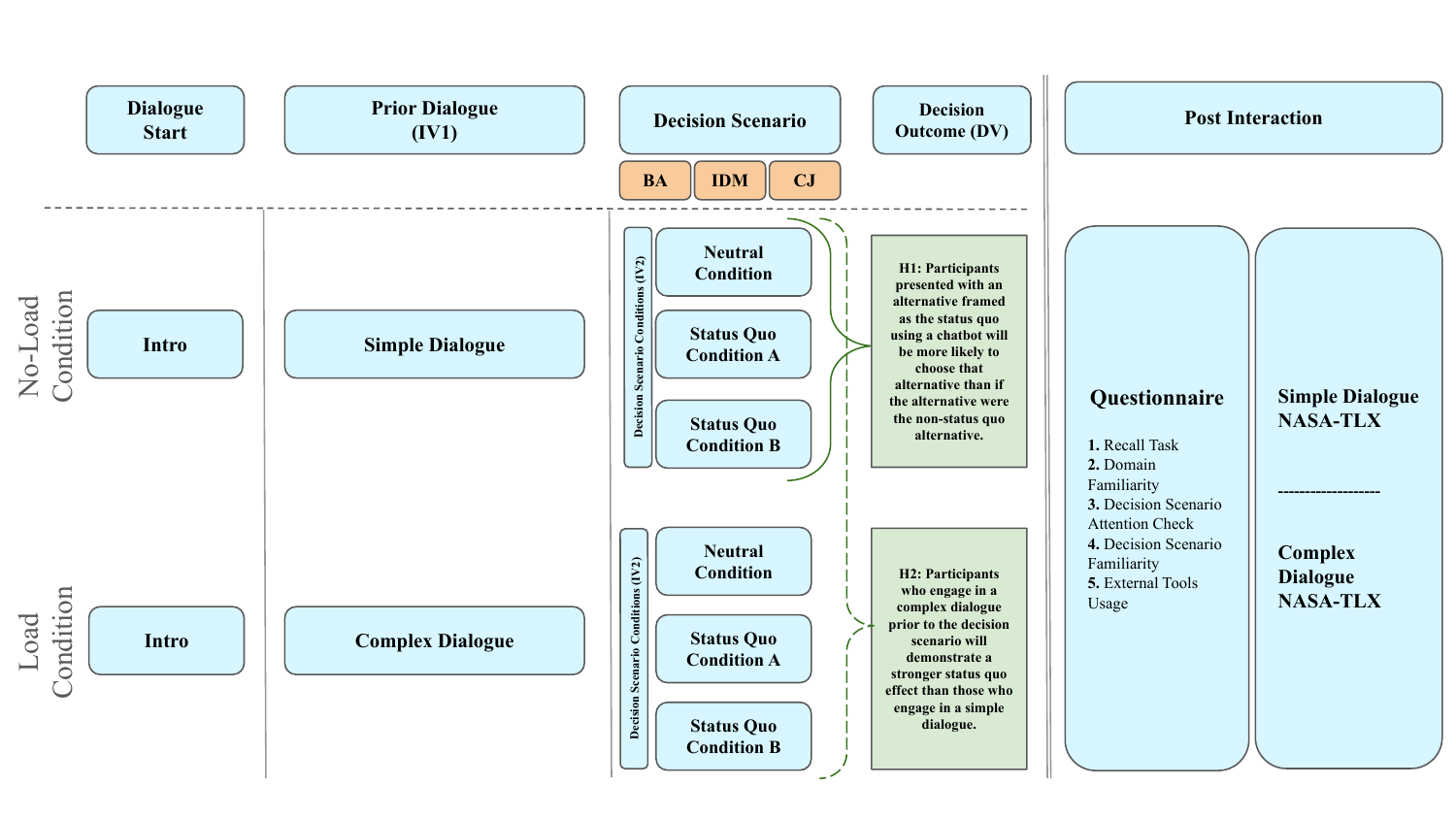}
    \caption[]{Overview of the experimental procedure and design. Task abbreviations: IDM — Investment Decision-Making, BA — Budget Allocation, CJ — College Jobs. IV1 and IV2 denote independent variables; DV indicates the dependent variable. }
    \label{fig:dox_h2}
    \Description{The figure presents a flowchart of the experimental procedure, illustrating parallel sequences for No-Load and Load conditions. In both conditions, participants begin with an introduction, then proceed to a dialogue task—either a Simple Dialogue in the No-Load condition or a Complex Dialogue in the Load condition. Following the dialogue, participants encounter a choice problem where they are randomly assigned to one of several sub-conditions (Neutral or Status Quo A/B) and then exposed to either a standard or alternative framing scenario. After making their choice, all participants complete a post-task questionnaire covering recall, domain familiarity, attention checks, scenario familiarity, and external tool usage, as well as the NASA-TLX workload assessment corresponding to their dialogue type. The flowchart depicts these steps and their organization across experimental conditions.}
\end{figure*}

\subsection{Design of the Experiments}

We employed a two-factor between-subjects design. The first independent variable (\textit{IV1}) is the \textit{prior dialogue complexity}, which manipulated the cognitive load experienced by participants during the dialogue. The second independent variable (\textit{IV2}) is the \textit{decision scenario conditions} detailed in Section ~\ref{sec:Method-Decision Scenarios and Experimental Conditions}.
We employed a $2 \times 3$ factorial design. Dialogue complexity had two levels (Simple-\textit{No-Load} vs. Complex-\textit{Load}), while the status quo factor had three levels (Neutral, Status Quo A, Status Quo B). 
The dependent variable (\textit{DV}) is participants' choices between alternatives in the decision scenarios. This is referred to as the \textit{decision outcome}. 
The \textit{moderator} variable is the \textit{perceived cognitive load} during the dialogue. We incorporated two indicators: the self-reporting \textit{NASA-TLX} and a \textit{Recall Task} as a behavioral indicator (See Appendix ~\ref{appendix:full_dox} for the details). The moderator variable allowed us to statistically test whether Complex dialogue resulted in higher mental demands on participants compared to Simple dialogue.
Decision outcomes were statistically compared across the three status quo conditions 
to test our first hypothesis (\textit{H1}), which posited that status quo framing alone, within a conversational context, would influence participants' choices.
To test our second hypothesis (\textit{H2}), we examined whether the magnitude of the status quo effect differed as a function of prior dialogue complexity (Simple vs. Complex), assessing whether increased cognitive load increased susceptibility to this bias. This approach allowed us to evaluate both the main effect of status quo framing and its potential moderation by dialogue complexity. To enhance methodological rigor, we incorporated additional design elements, including a decision scenario recall task to assess participant attentiveness and a domain familiarity measure to control for prior knowledge. Full details of these elements are provided in Appendix~\ref{appendix:full_dox}.

To ensure internal validity, we employed random assignment of participants to conditions and standardized all procedures across groups. Both the dialogue interactions and decision scenarios were fully scripted to eliminate potential confounding variables and ensure that only the intended factors varied between conditions. The domains of the prior dialogue (\textit{IV1}) and the decision scenario (\textit{IV2}) were intentionally distinct to prevent their interaction. Attention and manipulation checks were incorporated throughout the study to confirm participants' engagement and the effectiveness of our cognitive load manipulation. These steps collectively minimized alternative explanations for observed effects, allowing us to attribute differences in decision outcomes to our experimental manipulations.

To ensure ecological validity, both the prior dialogue tasks and the decision scenarios were designed to reflect realistic, domain-diverse interactions that participants might encounter in everyday life. The prior dialogue tasks spanned domains such as music, real estate, and banking, requiring participants to engage in natural, goal-oriented conversations that mirror common exchanges with conversational agents. The domains of these prior dialogues are adopted from the Schema-Guided Dialogue (SGD) dataset ~\citep{rastogi_towards_2020}. The decision scenarios similarly covered a variety of real-world decision-making contexts, including Budget Allocation (BA), Investment Decision Making (IDM), and College Job Selection (CJ). This diversity in domains enhances the ecological relevance of real-world human–agent interactions.

There is an inherent trade-off between internal and ecological validity in our design. To maximize internal validity, we intentionally selected unrelated domains for the prior dialogue and decision scenarios, thereby isolating cognitive load from domain familiarity. 
However, this separation can disrupt the natural conversational flow, making the chatbot interactions feel less authentic and potentially reducing ecological validity. 
However, such transitions are common in multi-goal dialogue systems, where users frequently engage in unrelated conversational goals~\citep{bernard_mg_2023}.
Additionally, the need for experimental control prevented open-ended dialogue, further limiting the realism of the dialogue. While these choices may slightly compromise the naturalness of the interaction, they are necessary to rigorously attribute observed effects to the experimental manipulations.

\subsection{Procedure}

Participants initiated the study by reviewing an information sheet that detailed the purpose, procedures, and ethical considerations of the research. After confirming their understanding of the study information, participants were directed to the experiment's homepage, where their Prolific ID was presented in an anonymous, irreversible format, alongside the consent form. To ensure explicit, informed consent, the consent checkbox was set to unchecked by default; participants were required to actively select ``Yes'' before proceeding to the main experiment.

\input{tables/new/table_st_sqnsq}

Upon providing consent, participants were assigned to one of the experimental conditions. As illustrated in Figure~\ref{fig:dox_h2}, the sequence of tasks varied depending on condition assignment. In the No-Load condition, participants first engaged in a Simple Dialogue with the agent, followed by a decision scenario. Conversely, those in the Load condition completed a Complex Dialogue before encountering the same decision scenario.

Following the dialogue and decision tasks, all participants completed a post-experiment questionnaire. This questionnaire included a memory recall task designed to assess participant attentiveness and engagement, as well as attention check questions to ensure data integrity. Additional survey items measured factors such as domain familiarity, perceived workload (using NASA-TLX), and use of external tools. By structuring the study in this way, with attention checks and recall tasks embedded throughout, the procedure maintained high participant engagement and data quality, enhancing the reliability of the findings.

\subsection{Power Analysis, Recruitment, and Data Integrity}
We conducted an \textit{a priori} power analysis using G*Power \citep{faul2009statistical}, targeting 0.80 power to detect a medium effect size ($\omega = 0.3$, $\alpha = 0.05$) following \citet{pancholi2009}. This required approximately 42 participants per condition (see Appendix~\ref{sec:appendix-power-analysis}). Participants (\textit{N}= 1256) were recruited via Prolific \citep{prolific2024}, compensated at \$8/hour. All procedures and hypotheses were preregistered on OSF:~\url{https://doi.org/10.17605/OSF.IO/PSXVF}.  
Data integrity was ensured via attention checks, recall tasks, and automated JSON logging. The anonymized dataset is available at~\url{https://doi.org/10.5281/zenodo.16541481}.

\subsection{Analysis Model Description}

To assess the effect of status‐quo framing on binary decision outcomes across scenarios, we fitted a generalised linear mixed-effects model (GLMM) using a binomial logistic regression. The dependent variable was the participant's choice between two alternatives; the key predictor was the Status Quo Condition (option A or B framed as status quo). Covariates included domain familiarity, attentiveness, AgeBracket, country, and sex. A random intercept for the dialogue domain accounted for between‐domain variability. Models were estimated in R's lme4 package~\citep{r_lme4} with a logit link.

\begin{align}
\nonumber 
\text{Decision Outcome} \sim\ 
& \text{Status Quo Condition} + \\\nonumber 
& \text{Prior Dialogue Familiarity} + \\\nonumber 
& \text{Prior Dialogue Attention} + \text{Age} + \\\nonumber 
& \text{Sex} + \text{Country Of Residence} + \\\nonumber 
& (1 \mid \text{Prior Dialogue Domain})\nonumber 
\end{align}

We applied this model separately to the No-Load and Load conditions to examine the effect of status quo framing under different prior dialogue conditions. Additionally, the same analytical approach was consistently used across all three decision scenarios. The data cleaning and dataset description are available in the Appendix ~\ref{appendix:demographics}.

\begin{figure*}[htpb]
    \centering
    \includegraphics[width=1.0\linewidth]{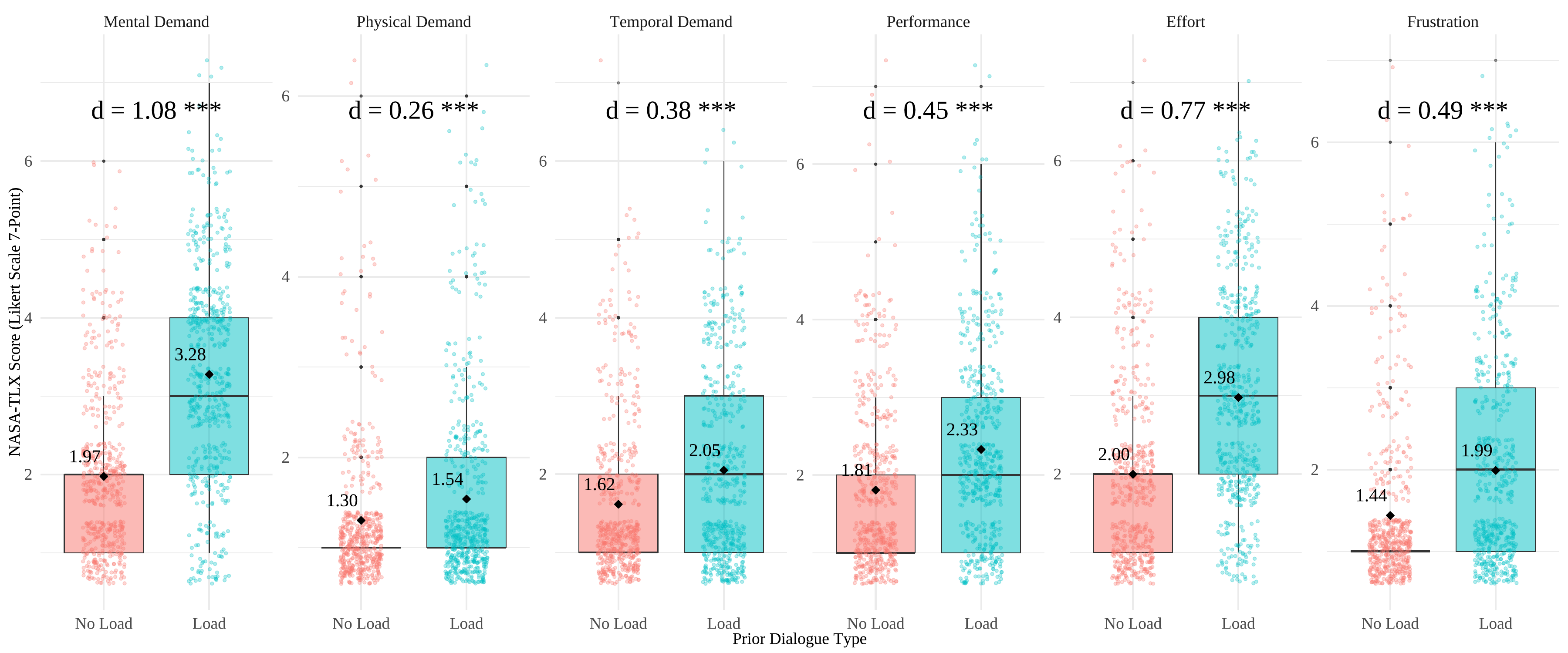}
    \caption[]{NASA-TLX scores show significantly higher perceived Mental demand and Effort under the Complex Dialogue condition, confirming the effectiveness of the cognitive load manipulation.}.
    \Description{The figure presents a set of boxplots for NASA-TLX workload scores for status quo bias, comparing No Load and Load (complex dialogue) conditions. Six workload dimensions are plotted, with individual scores, means, boxplot elements, and annotated effect sizes and significance values. For instance, Mental Demand averages 1.97 in the No Load condition and 3.28 in the Load condition (d = 1.08). Effort is 2.00 for No Load and 2.98 for Load. The x-axis labels the conditions (No Load, Load), while the y-axis represents the NASA-TLX Likert scale.}
    \label{fig:ntlx}
\end{figure*}

\section{Results}
This section presents the results of our empirical study. To maintain focus on the data, we report the findings without interpretation or theoretical analysis. A detailed discussion and interpretation of these results is provided in the Discussion section. 

\subsection{Status Quo Effects in Human–Chatbot Interactions}

This subsection examines whether status quo framing, when delivered through a chatbot, influences user decision-making across different scenarios in the absence of cognitive load.

\subsubsection{How to Read the Table.}

Table \ref{tab:ST-SQNSQ} reports the proportions of participants choosing each option under status-quo (SQ), neutral (N), and non-status-quo (NSQ) framings. Following ~\citet{samuelson_status_1988} and ~\citet{xiao_revisiting_2021}, a status quo effect is inferred if $SQ > N > NSQ$, indicating that designating an option as the status quo raises its selection rate above baseline (N) and that removing that designation lowers it.

\subsubsection{Interpretation of Results.}

In the Budget Allocation (BA) scenario, framing 60A40H as the status quo increased its selection from 14\% to 51\% ($h = 0.79$, $p < .001$) with a large effect size, closely mirroring the magnitude reported in earlier studies ($h = 0.78$, $p < 0.025$)~\citep{samuelson_status_1988}. This suggests that even in a chatbot-mediated context, subtle framing cues significantly bias decision outcomes. Similarly, in the College Job (CJ) scenario, framing College A as the status quo increased its selection from 53\% to 70\% ($h = 0.47$, $p < .05$), reflecting a medium status quo effect. The original study reported a large effect size ($h = 1.26$, $p < .001$)

In contrast, the Investment Portfolio (IDM) scenario did not exhibit a statistically significant status quo effect in our study. Selection rates for the moderate-risk and high-risk options remained relatively consistent across framing conditions, with a non-significant result ($h = 0.09$, $p = .657$) and negligible effect size. This indicates that participants' decisions in this scenario were not influenced by status quo framing. 
However, it is worth noting that the original study ~\citep{samuelson_status_1988} reported a marginal effect ($h = 0.38$, $p = .069$) on this scenario.

Overall, these results show that chatbot-mediated status-quo framing significantly biases decisions, reliably triggering status quo effects with varying strength across decision scenarios. In short, the \textit{IV2} influenced \textit{DV}, providing support for hypothesis \textit{H1}. Importantly, this work contributes to the empirical establishment of a human baseline for status quo effects within dialogue-based interactions.

\begin{figure*}[htpb]
    \centering
    \includegraphics[width=1.0\linewidth]{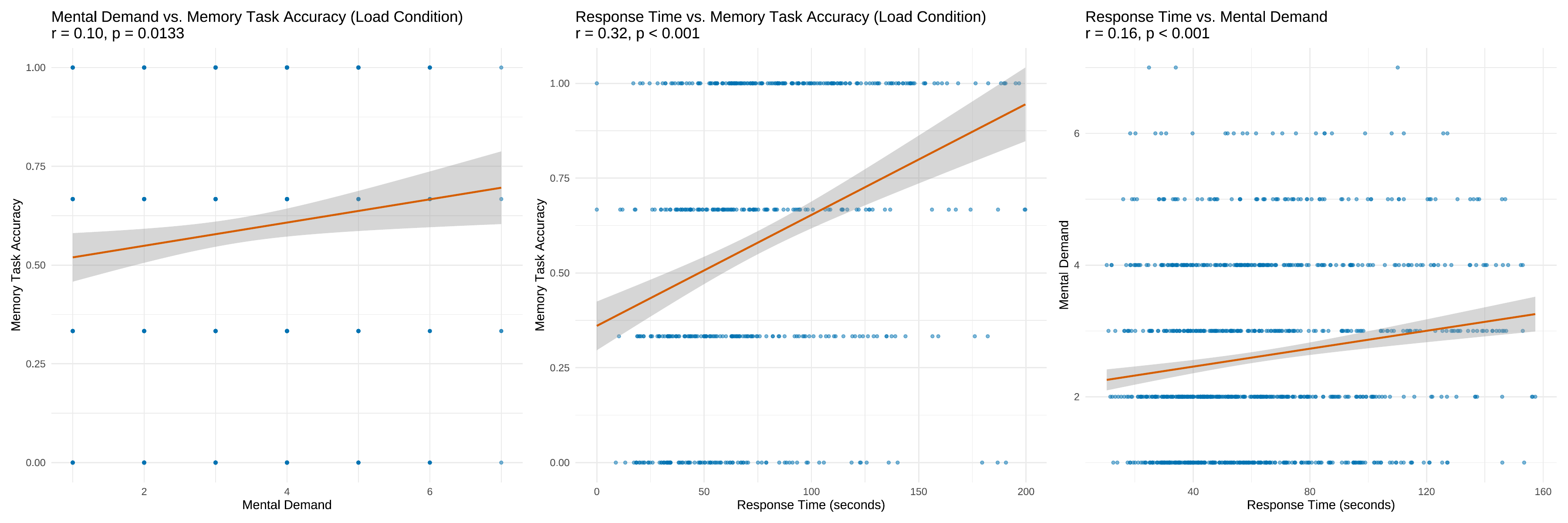}
    \caption[]{Scatter-plots with regression lines showing associations between Mental Demand and Memory Task Accuracy (left), Response Time and Memory Task Accuracy (center), and Response Time and Mental Demand (right), the first two under Load condition. Shaded bands represent 95\% confidence intervals.}
    \Description{The figure presents three scatter-plots, each overlaid with a linear regression line and 95\% confidence interval bands. The left plot shows the relationship between Mental Demand (x-axis) and Memory Task Accuracy (y-axis) under the Load condition, with the Pearson correlation coefficient (r = 0.10, p = 0.0133) indicated at the top. The center plot depicts Response Time (seconds) on the x-axis and Memory Task Accuracy on the y-axis, also under the Load condition, with a reported correlation of r = 0.32 (p < 0.001). The right plot displays Response Time (x-axis) against Mental Demand (y-axis), with a correlation of r = 0.16 (p < 0.001). Each blue dot represents a participant's data point. The regression lines are shown in orange, with shaded regions indicating the 95\% confidence intervals for the fit.}

    \label{fig:correls}

\end{figure*}

\subsection{Experiment Two: Influence of Prior Dialogue Complexity on Subsequent Decision-Making}
We tested Hypothesis 2 in two steps: (1) verifying that prior dialogue complexity induces cognitive load, which is measured by NASA-TLX self-reports and performance indicators, and (2) assessing its interaction with subsequent decision-making.

\subsubsection{Perceived Cognitive Load}
We compared NASA-TLX scores and performance metrics between  No-Load and Load conditions to confirm that complex prior dialogue generates cognitive load in chatbot interactions.

Figure~\ref{fig:ntlx} presents NASA-TLX scores across six dimensions for No-Load (Simple) versus Load (Complex) conditions, including means, Cohen's d, and significance markers ($*** p < .001$). 
Mental Demand increased significantly from $M = 1.97$ to $3.28$ ($d = 1.08$, $p < .001$). Similarly, Effort also increased from $M = 2.00$ to $2.98$ ($d = 0.77$, $p < .001$), indicating that the arithmetic, memory, and the length of the dialogue contributed to the perceived cognitive load, respectively.
Performance, Frustration, and Temporal Demand also rose significantly, however, with small to medium effect sizes, while Physical Demand showed a minimal effect ($d = 0.26$). These results confirm that complex prior dialogue in chatbot interactions substantially increases perceived cognitive load, specifically in Mental Demand and Effort.

\input{tables/new/table_ct_sqnsq}

We assessed the alignment of self-reported and behavioral indicators of cognitive load under the Complex Dialogue condition. Figure \ref{fig:correls} shows the correlations among recall accuracy, decision response time, and Mental Demand.
Behavioral indicators further validated our cognitive load manipulation. Recall accuracy correlated positively with Mental Demand ($r = 0.105$, $p = .013$), showing that participants who remembered memory task details also reported higher workload. Response time and recall accuracy correlated positively ($r = 0.318$, $p < .001$), indicating that participants who recalled paid additional attention to the choice problem.
Moreover, participants in the Complex Dialogue took significantly longer to respond than in the Simple Dialogue ($d = 0.59$, $p < .001$), confirming that increased prior dialogue complexity increased both perceived and measured cognitive load. 
These converging findings validate our manipulation of cognitive load via prior dialogue.

\subsubsection{Decision-Making Under Complex Dialogue}

Building on prior results, we examined whether cognitive load from complex prior dialogue interacts with status quo effects in three scenarios (Table \ref{tab:CT-SQNSQ}).
Budget Allocation under Load still showed a strong status quo effect: presenting 60A40H as default raised its choice rate from 16\% to 47\%. Similarly, when 50A50H was made status quo, the choice rates increased from 53\% to 84\%, confirming a significant status quo effect ($h = 0.81$, $p = .001$).
Investment Portfolios showed no status quo effect: choice rates were unchanged by presentation, indicating resistance to status-quo presentation, regardless of load ($h = 0.04$, $p = .207$).

College Jobs under Load exhibited a moderate–strong effect: presenting College A as status quo resulted in choice rates increasing from 57\% to 71\%, similarly for College B, which increased from 29\% to 43\%, indicating a significantly moderate status quo effect($h = 0.59$, $p = .001$). This effect was slightly stronger than under No-Load.
In summary, cognitive load did not alter the presence, but increased the strength slightly of status quo effects in Budget Allocation and College Jobs; however, it is negligible. However, the Investment Portfolios remained unaffected by the status quo presentation and cognitive load manipulation.

This human experiment investigated whether contextual cognitive load interacts with biased decision-making behavior in dialogue-facilitated settings. 
\textit{IV1} had a significant effect on the moderator variable, cognitive load, indicating that our manipulation was successful in altering cognitive load levels across conditions.
However, the dependent variable (\textit{DV}) was not significantly influenced by the interaction between \textit{IV1} and \textit{IV2}, leading us to reject hypothesis \textit{H2}. No evidence was found to support $H2$; therefore, we conclude that Status quo bias was not influenced by the prior dialogue complexity. 

\subsubsection{Other Factors and Random Effects.}
We included covariates in our statistical model: Prior Dialogue Familiarity, Prior Dialogue Attention, Age, Sex, and Country of Residence, to control for potential confounding effects. Additionally, a random intercept was included for the Prior Dialogue Domain to account for domain-level variance. However, none of these variables showed a significant influence on the decision outcomes across conditions, indicating that the observed effects were robust to demographic and contextual variation. The results of these factors are detailed in Appendix~\ref{sec:appendix-model_results}.

\section{Agent Experiments}

We replicated the experimental procedure used with human participants with LLMs. Each agent was constructed using two key inputs: (a) the demographic attributes available from Prolific (e.g., age, gender, education, and country of residence) and (b) the transcript of the participant's dialogue up to the decision scenario or choice problem; this chat includes the prior dialogue, as shown in the Chat Transcript. Agents presented with the same decision scenarios as their corresponding human participants and were asked to act as participants and predict the decision the participant would make. This design ensured that the information available to the agents mirrored the information grounding human decisions.
We then compare the LLM agent's simulated response to the corresponding participant's response to measure predictive accuracy and analyze whether the agent is leveraging the individual-specific cues. 
Further, we analyze LLM agents’ simulated responses collectively to see if they can reproduce patterns of cognitive bias at the sample level.
By comparing the distribution of simulated and human responses, we assess how accurately LLMs represent biased human decision-making behavior. Further, we do this across three different levels to determine how much explicit instruction is needed for alignment.

\input{tables/h3_h1}
\subsection{Human-Likeness Prompts}

A central design choice in the agent experiments was how to instruct LLMs to act as human participants. Following recent work on LLM behavioral prompting ~\cite{binz2024turning,liu2025rationality, ying2025benchmarking}, we adopted a series of human-likeness prompts that varied in the degree to which the agent was instructed to emulate human reasoning. Prior research suggests that LLMs may default to more rational or normatively consistent behavior than humans ~\citep{liu2025rationality}, raising the concern that observed biases in classic decision scenarios may reflect statistical pattern matching rather than genuinely human-like behavior. To test this, we systematically varied the level of human-likeness in the prompts, ranging from minimal framing to explicit instructions to behave in a bias-susceptible manner.
\input{others/chat_transcript}
We implemented three levels of human-likeness. At Level 1, agents received only a minimal role instruction: ``You are a participant in a research study.'' This established an experimental setting without explicit guidance on how to respond, allowing us to assess the agent's baseline behavior. 
At Level 2, agents were encouraged to simulate more naturalistic human responses with the prompt: ``You are a human participant in a research study. Please answer questions as naturally as you would in everyday life.'' This formulation aimed to elicit more ecologically valid answers while avoiding explicit mention of biases. 
At Level 3, we explicitly instructed the agents to act as humans prone to cognitive biases: ``You are a human participant in a research study. Therefore, act as a human. Be highly susceptible to cognitive biases such as framing, status quo bias, and anchoring when reasoning and answering questions. Avoid overthinking and lean into intuitive, sometimes irrational judgments.''

\subsection{Technical Details}
All agent simulations were conducted using the OpenAI API with GPT-4.1, GPT-4.1-mini, GPT-5, and GPT-5-mini via chat. completions.endpoint. To ensure reproducibility, we used batch mode, fixed the random $seed$ to 42, set the $temperature = 0$ for deterministic outputs, and logged the system fingerprint returned by the API to track  
~\citet{openai_advanced_usage} model versions. Annotation of the output was conducted separately using three different LLMs: GPT-4.1, GPT-4.1-mini, and GPT-5-mini. Inter-rater reliability (IRR) was calculated to choose the annotation for analysis.

\subsection{Findings}

\subsubsection{LLM Prediction Accuracy Across Human Likeness Levels}
\label{sec pred}

\input{tables/ind_pred}

We evaluated how precisely LLMs predicted individual human decisions given the prior dialogue transcript under varying levels of human-likeness prompting (HL1 to HL3). 
As shown in Table~\ref{tab:individual-hl}, GPT-4.1 achieved the highest precision across all three levels, peaking at 0.685 under HL2. This is significantly higher than the random chance. Its smaller variant, GPT-4.1-mini, also performed competitively, particularly at HL2 (0.655) and HL1 (0.635), though with a noticeable drop at HL3. Interestingly, GPT-5 exhibited increasing precision across the human-likeness levels, reaching 0.615 at HL3. In contrast, GPT-5-mini showed the lowest precision overall, though its performance improved steadily from 0.456 at HL1 to 0.545 at HL3. Precision, recall, and F1 scores for the simulated and human responses are reported for each decision scenario in detail in the Appendix ~\ref{indpredall}.

\subsubsection{Status Quo effect observed in human experiments reproduced at the sample level}

Table~\ref{tab:h3_h1} presents Cohen's h values to compare both human and model behaviors across three decision scenarios under varying levels of human-likeness. 
Most conditions demonstrated clear evidence of bias in both Simple and Complex dialogue for human participants. Investment Decision Making (IDM) showed only a marginal effect similar to the original study, where our human experiments showed similar outcomes. 

\input{tables/precision}

Agents displayed varying levels of bias across the three human-likeness conditions. When explicitly instructed to behave in a biased manner (HL3), agents showed strong bias across all decision scenarios, including Investment Decision Making (IDM), where no significant bias was observed in the human experiments. 
This resulted in false positives; that is, when human experiments showed no significant bias, the LLM agents, while simulating the HL3 condition, predicted a strong bias, revealing a forced, biased behavior when explicitly asked. 
Consequently, the precision, the proportion of cases where bias was reproduced in simulation that actually matched the bias observed in human experiments, was only 0.67 in HL3.
In contrast, when agents were given a more neutral prompt, the HL1 and HL2, precision improved to 1.0 as show in Table ~\ref{tab:confusion_matrices}. HL1 and HL2 faithfully reproduce the bias as they closely align with human experiment results. These findings act as preliminary evidence for our hypothesis \textit{H3} that LLMs can accurately emulate or represent biased human behavior in simulations.

\input{tables/h3_h2}

\subsubsection{LLMs reproduce observed biased human behavior under complex prior dialogue.}

To test Hypothesis 2 with LLMs, that cognitive load from prior dialogue interacts with the Status Quo effect, we conducted independent t-tests comparing effect sizes between Simple and Complex dialogue conditions.

To investigate whether LLMs could capture this interaction pattern, we examined the direction and magnitude of effect size changes across dialogue conditions using z-scores. 
For example, in College Jobs, the effect size and the confidence intervals for humans ranged from small to medium ($h=0.46$, $p < .01$), and for complex dialogue it ranged from medium to large ($h=0.58$, $p < .01$) (as shown in Table ~\ref{tab:h3_h1}), resulting in a positive z-score of 0.88 (as shown in Table ~\ref{tab:h3_h2}), indicating a relatively increase in bias under cognitive load. 
However, in Human-Likenesss 1 (HL1), the LLM's effect size remained unchanged between conditions, producing a negative z-score of -0.02, meaning the LLM didn't observe any change. 
Interestingly, HL3 showed a positive z-score of 3.32, indicating a pattern closer to human responses, where bias increased under complex dialogue. Similar trends were observed for the Budget Allocation and Investment scenarios, where humans showed no change in direction, which was accurately mirrored in HL3 but not HL1 or HL2.
We calculated the Mean Absolute Difference (MD) between the human z-scores and those of each human-likeness condition. Quantitatively, HL3 is the closest to Human performance (MD = 0.92), while HL1 (MD = 2.19) and HL2 (MD = 1.47) (Lower MD is better) showed a poor match. 

\subsection{Analysis Across Models}

Sample-level precision shows how often the agent correctly reproduces human biases without generating false positives (Number of cases predicted correctly divided by the total number of cases predicted correctly and incorrectly). A higher precision means that the bias observed in simulated responses matches human experiments, while lower precision indicates that LLMs tend to see bias where humans do not exhibit. Mean Absolute Difference, on the other hand, reflects whether the agent captures the pattern of change in biases, such as increases or decreases, under complex dialogue. An MD close to zero suggests the LLM's behavior aligns with bias dynamics observed in human experiments while interacting with cognitive load, whereas a higher MD indicates a mismatch.

\input{tables/precision_and_mad}

Table~\ref{tab:precision_mad_all_models} summarizes precision and mean absolute difference (MD) scores for each model across the three Human Likeness levels.
Among all models, GPT-4.1-mini consistently demonstrated the strongest performance. Its precision was perfect for HL1 and HL2 (1.000 each) and remained relatively high for HL3 (0.667). It also achieved the lowest MD across HL1 and HL2 (2.793 and 0.883, respectively), indicating that LLM responses were close to human responses.
GPT-4.1 also performed well, with perfect precision on HL1 and HL2 and moderate precision on HL3 (0.667), along with low MD values (0.923). This suggests that both GPT-4.1 models reproduced human biases accurately.
By contrast, the GPT-5 family performed less effectively. GPT-5-mini achieved perfect precision on HL1 and HL2 but dropped to 0.667 for HL3, with higher MD values (2.863–2.430), reflecting larger deviations from human patterns. GPT-5 showed the weakest results: its precision was low across all conditions (0.000–0.667), and while its MD scores were numerically smaller in some cases, this reflects inconsistent alignment with humans rather than genuine closeness.
Taken together, these results highlight a clear divide between model families. The GPT4.1 models, particularly GPT4.1-mini, captured both the biased human behavior and prior dialogue complexity interactions with a certain degree of certainty compared to the GPT5 family.

\subsection{Ablation \& Perturbation}

To better understand which components of our experimental setup contributed to the LLM's ability to reproduce human-like decision-making behavior, we conducted a series of ablation studies. These ablations systematically removed or isolated different parts of the input, such as demographics, human-likeness prompts, prior dialogue components (arithmetic and memory), to identify what elements were essential for reproduction of bias and behavior in a complex dialogue setting.

\input{tables/ablation}

We began by testing a minimalist setup that presented only the decision scenario to the model, excluding demographic information, prior dialogue, and human-likeness prompts. In this condition, both GPT-4.1 and GPT-4.1-mini still produced biased responses, suggesting that bias is a stable feature of LLM behavior. However, the alignment of effect sizes with complex conditions differed. GPT-4.1 achieved high precision (0.80) but large deviations from human effect sizes (MD = 4.957, RMSE = 6.939), while GPT-4.1-mini showed lower precision (0.500) but much smaller errors (MD = 1.037, RMSE = 1.057). This suggests that while both models replicated baseline biases, they failed to capture how dialogue complexity shapes them, reinforcing that prior dialogue is necessary for reproducing human-like interaction effects. More details of the models for each human likeness are presented in Table~\ref{tab:ablation_table_new}.

Next, we isolated the memory component of prior dialogue and excluded the rest, where models were asked to retain specific information before responding to the decision scenario. GPT-4.1 maintained similar precision (0.80) with moderate error (MD = 5.800, RMSE = 5.932), while GPT-4.1-mini achieved perfect precision (1.0) but showed larger deviations from human responses (MD = 6.183, RMSE = 8.803). These results suggest that including memory cues increases the likelihood of a biased response. This suggests that the memory component could play a role, but it is not sufficient.

We then examined the effect of the arithmetic component of prior dialogue, removing memory but retaining numerical comparison tasks. In this condition, both models achieved perfect precision (1.0), indicating strong bias reproduction. However, GPT-4.1 aligned much more closely with human data in terms of effect size (MD = 1.10, RMSE = 1.104), while GPT-4.1-mini exhibited larger errors (MD = 4.137, RMSE = 4.577). This indicates that arithmetic reasoning alone was not sufficient to emulate interaction between dialogue complexity and bias. Overall, the ablation studies show that prior dialogue arithmetic and memory components are not mutually exclusive in contributing to driving alignment between LLM behavior and human biases while simulating interaction between dialogue complexity and bias.

To further examine whether LLMs rely on human responses in the prior dialogue when predicting individual decisions, we conducted a human response perturbation analysis. Specifically, we replaced the human responses in the chat transcripts(~\ref{DialogueTranscripts}) with randomly generated text while keeping all other aspects of the prompt unchanged. We observed that the precision of the LLMs' predictions dropped only slightly compared to when the original human responses were included, as shown in Table ~\ref{tab:individual-hl}. This suggests that the models may not be meaningfully leveraging cues from the human's dialogue behavior, and instead rely more heavily on decision scenarios and prior dialogue to make their predictions.

\section{Discussion}

This study explored whether large language models (LLMs) can represent biased human decision-making behavior, particularly cognitive biases like the Status quo effect, in a simulated conversational context. We investigated two key questions: (1) whether they can predict decision-making at the individual level with demographic, limited prior dialogue, and (2) whether LLMs can reproduce biased decision-making at the sample level.

\subsection{Individual-Level Prediction}

While LLMs performed reasonably well in predicting individual decisions (e.g., ~68\% precision), perturbation tests showed that these predictions did not depend heavily on the actual participant utterances in the dialogue. Replacing participant utterances with random text resulted in only minor drops in precision. 
These findings indicate that LLMs primarily rely on general patterns in the dialogue task and decision scenarios, rather than fine-grained, individual-specific cues. This may be due to the absence of explicit personalization or the relatively short length of conversations, which results in sparse individual-level cues. Consequently, the model's capacity for individual-level prediction remains limited. 
These results provide evidence addressing the question posed earlier in the introduction, which is, whether the biased behavior exhibited by LLMs arises from individual-specific cues present in the prior dialogue.

\subsection{Sample-Level Bias Reproduction}

Our findings demonstrate that LLMs are capable of reproducing the Status Quo bias at the sample level, but this ability is highly dependent on how the models are prompted. 
In the human study, participants consistently showed a preference for options framed as the status quo when making decisions through a chatbot interface (H1).
When evaluated under HL1 and HL2 prompting conditions, where LLMs were instructed to act like humans or imitate human responses, the models successfully replicated this behavioral pattern with high precision.
When prompted to act with biases (HL3), LLMs overestimated the bias, resulting in false positives and lower precision. 
Alignment overfitting refers to the phenomenon where language models become excessively responsive to certain prompts, often ``over-aligning'' their outputs to perceived instructions or expectations ~\citep{wolfalignment2024, miehling-etal-2025-evaluating}. 
In our case, we can observe that the model's increased sensitivity to HL3 resulted in exaggerated display bias, rather than a faithful emulation of human decision patterns.

Overall, LLMs can reproduce sample-level tendencies in decision-making, particularly for well-documented effects like the Status Quo bias. 
However, while this alignment at the sample level is promising, it raises important questions about whether the observed behavior reflects deeper cognitive processes or merely statistical associations 
based on the extensive presence of similar decision tasks in their training data.

The results from the individual level and sample level, combined, raise important questions about the depth at which LLMs can represent biased human decision-making behavior. While they can match average human behavior at the sample level, they struggle to reason about individual-specific decision tendencies. We suspect this due to the sparse cues in the participants' responses in the dialogue. In other words, for our task, the model understood ``what humans tend to do,'' but not ``what this human is likely to do.''

~\subsection{Reproducing Contextual Interactions}

Our findings contribute to ongoing discussions around whether LLMs can represent biased human decision-making behavior that can reproduce human cognitive biases in a simulated decision-making dialogue. 
While recent work has suggested that LLMs may behave more rationally than humans~\citep{liu2025rationality}, our study under HL1 and HL2 prompting conditions shows that LLMs can reproduce status quo bias with high precision. However, this alone does not confirm that LLMs are simulating human cognitive processes. It remains possible that these models are simply matching patterns based on learned statistical associations, especially given the widespread use of status quo decision scenarios in existing datasets. If LLM behavior arises from such surface-level pattern recognition rather than emulating underlying cognitive mechanisms, its use in behavioral simulation may be misleading. In these cases, LLMs risk producing superficial or inaccurate representations of human decision-making, limiting their reliability for simulation.

Cognitive biases often interact with contextual factors such as cognitive load, as predicted by dual-process theory~\citep{kahneman_thinking_2011}. However, this interaction is not consistent across all types of biases. In our experiment focusing on the Status Quo bias, we found that increased prior dialogue complexity consistently did not affect observed bias across all decision scenarios in human participants.
In most cases in HL1 and HL2, where the models were instructed to act like humans but without explicit reference to cognitive biases, LLMs failed to align with human behavior under interactions. 
In fact, their responses under complex dialogue conditions showed effect sizes that diverged from human behavior (For instance, GPT4.1-mini budget allocation in HL1, GPT4.1 budget allocation in HL2, and many more). The effect sizes and overlaps can be observed from the forest plot in Appendix ~\ref{appendix:all_model_forestplot}. In contrast, under HL3, where the LLMs were explicitly instructed to act as humans susceptible to cognitive biases, the models showed their alignment; however, this is because they overestimated the bias in all cases. This is not desirable.

However, the effect sizes and confidence intervals in gpt-4.1-mini in HL2 leave an interesting note. The interactions converging toward human behavior suggest that LLMs may be capable of simulating context-sensitive bias patterns when explicitly prompted. However, this is only preliminary evidence, leaving open the possibility that LLM-generated behavior could more closely resemble human patterns and offer potential for more realistic behavioral simulation. Further investigation into a broader range of biases is required for stronger generalization.

\section{Conclusion}

Our experiments show that LLMs are capable of representing human biased behavior that is susceptible to status quo bias within simulated dialogue. Notably, LLMs exhibit this behavior with minimal prompting and without relying on individual-specific cues. In addition, evidence suggests that LLMs not only represent human biased-behavior but also interact with contextual factors such as dialogue complexity.
However, this is preliminary evidence, shedding some light on the faithful emulation of human biased behavior, and requires further investigation into other biases where contextual factors interact with the bias. 

LLM Simulation ~\cite{hwang2025human} is an emerging field that can benefit from our findings. 
While prior works in this area have shown that LLMs can replicate survey responses or general attitudes ~\cite{park2024generative}, our study demonstrates that LLMs can reproduce biases with precision with minimal prompting. This further sheds some light on the behavioral fidelity that is critical for simulations of real-world decision-making. 
Simulations that incorporate only surface-level biases may fail to capture how those biases vary depending on dialogue structure, memory demands, or task complexity. Our work provides preliminary evidence to suggest that LLMs can potentially serve as more realistic proxies for humans.
Moreover, we see a design opportunity for adaptive, bias-aware systems that adjust dialogue strategies in real time, simplifying interactions under high cognitive load or transparently presenting defaults when bias effects may unduly influence users.

\section*{Limitations}

The study in this paper uses classic decision scenarios that are abstract and text-based. While this approach allows for experimental control, it may not fully capture the complexity of real decision-support dialogues found in domains like finance, recruitment, or healthcare. This implies that our findings may not generalize to practical, real-world settings. We aim to apply this framework to more realistic, domain-specific conversations to improve external validity.

This study specifically investigates the Status Quo bias, providing a tightly controlled and replicable foundation for conversational agents and cognitive bias research. 
However, because different cognitive biases (e.g., anchoring, framing, confirmation) may operate through distinct psychological mechanisms, these findings should not be assumed to generalize beyond Status Quo effects without further targeted research.

This work does not seek to characterize or analyze various sources of dialogue complexity. Rather, our focus is on the case where such complexity is already present and how the resulting cognitive load influences susceptibility to status quo bias. Different dialogue strategies, such as open-ended versus closed questions, or conservative versus liberal dialog style, may also contribute to dialogue complexity~\citep{brachten2020ability}. However, a detailed analysis of these styles and the naturalness of the dialogue is beyond the scope of this paper.

The study relies on self-reported (NASA-TLX) and behavioral (recall accuracy, response time) indicators for cognitive load. While these are widely accepted, they do not capture real-time cognitive load. Incorporating physiological measures (e.g., eye tracking, EEG, or pupil dilation) ~\citep{Haapalainen2010} in future studies would offer more precise and dynamic assessments of cognitive load during dialogue.

Our agent experiments were conducted exclusively using OpenAI's GPT-4.1 and GPT-4.1-mini models. Although our approach is model-agnostic and can be easily extended to other LLMs (e.g., Claude, Gemini, Mistral), we leave this as future work, as this is not the focus of the paper. However, some comparison across the generations was reported in this paper. The released code base is designed to be modular and supports replication with alternative models, enabling broader benchmarking across LLMs.

Despite implementing multiple safeguards to ensure data quality, such as memory recall tasks, attention checks, and response time analysis, there remains a possibility of data contamination. We found no clear evidence of LLM-assisted responses (e.g., based on response similarity, length, or timing patterns). Appendix ~\ref{appendix-dialogue-validation} provides a detailed account of participants' dialogue validation. However, the risk cannot be entirely dismissed. 

Although this study focuses on dialogue, the same approach can be extended to other settings. For example, an LLM could be asked to perform decision-making in visual interfaces, where a default checkbox might act as the bias-inducing element, and nearby text might provide contextual information. Still, dialogue remains the most natural and effective way to study how bias and cognitive load interact in human–agent dialogue.

\section*{Statement of Ethics}
All human-subject data were collected under ethical approval, anon-ymized prior to analysis, 
and handled in accordance with data protection guidelines; no personally identifiable demographic information, such as participant names or Prolific IDs, was given as input to the LLMs.
This study was approved by the Institutional Review Board (IRB) and adheres to the ethical guidelines established by the institution. Details have been elided for anonymous review, can be provided on request. 
\textit{(HREC-LS): LS-C-25-001-Pilli-Nallur}.

\begin{acks}
This work was conducted with the financial support of the Research Ireland Centre for Research Training in Digitally-Enhanced Reality (d-real) under Grant No. 18/CRT/6224. 
For the purpose of Open Access, the author has applied a CC BY public copyright licence to any Author Accepted Manuscript version arising from this submission.

\end{acks}

\bibliographystyle{ACM-Reference-Format}
\bibliography{bib}

\appendix

\input{tables/supp_material/all_decision_scenarios}
\input{others/text_adj_sqb}

\newpage
\input{tables/supp_material/simple_task_questions}

\input{tables/supp_material/complex_task_questions}
\newpage
\input{others/Additional_DOX}
\newpage
\input{others/validation}

\newpage
\input{tables/regression_model_results}

\newpage
\section{Sample Planning and Data Collection}
\label{appendix:sample_planning_and_data_collection}
\input{others/sample_planning}


\newpage
\section{Dataset Description}
\label{appendix:demographics}
\subsection{Demographics}
\input{tables/demographics}
\subsection{Data Cleaning}

\begin{itemize}
    \item Data cleaning excluded:
    \begin{itemize}
        \item 19 participants with ``Invalid'' outcomes
        \item 40 participants who failed to recall the decision scenario
        \item 49 participants familiar with the decision scenarios
        \item 63 participants who used external tools
    \end{itemize}
    \item Final sample: \(N = 1100\)
    \item Mean age: 41.5 years (SD = 13.3; range: 18–87; median = 39.5)
    \item Gender distribution:
    \begin{itemize}
        \item 50.5\% female (n = 556)
        \item 49.5\% male (n = 544)
    \end{itemize}
    \item Most common countries of residence:
    \begin{itemize}
        \item United Kingdom (n = 778)
        \item United States (n = 308)
        \item Ireland (n = 14)
    \end{itemize}
\end{itemize}

\newpage

\section{Forest Plot}
\label{appendix:all_model_forestplot}

\begin{figure*}[htpb]
    \centering
    \rotatebox{90}{\includegraphics[scale=0.2475]{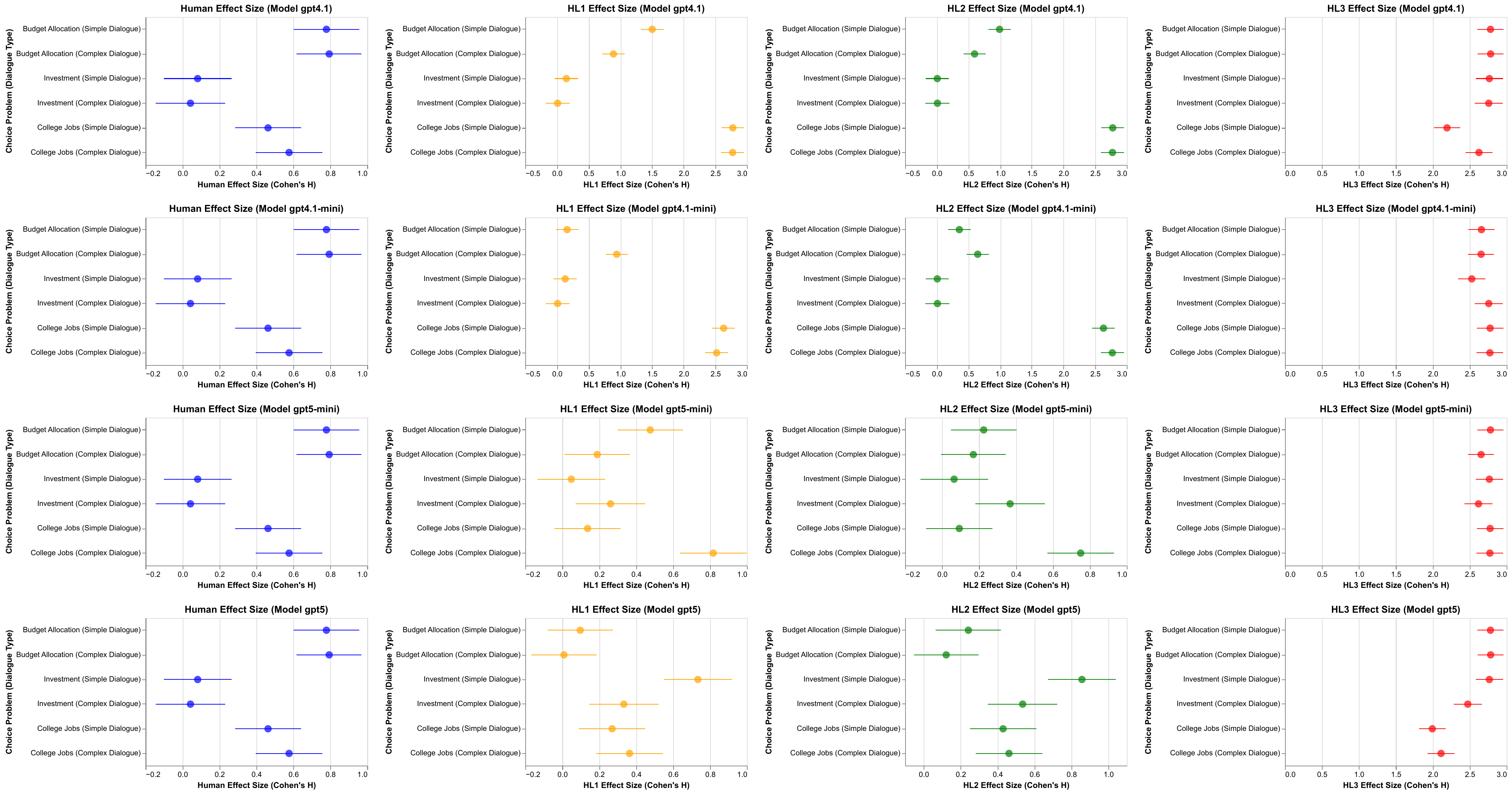}}
    \caption[]{Forest plot for effect sizes and 95\% confidence intervals for all models compared with Human experiments.}
   \Description{Forest plot arranged in a grid showing effect sizes with 95\% confidence intervals for multiple dialogue conditions. Columns represent different decision scenarios, and rows represent different sources: GPT-4.1, GPT-4.1-mini, GPT-5, and GPT-5-mini, and human participants. Each panel contains colored points with horizontal lines indicating confidence intervals. The y-axis shows standardized effect size (Cohen's h), and the x-axis lists dialogue conditions. Colors distinguish sources, with blue representing humans and red, green, and orange representing different language models.}
    \label{fig:forestplot}
\end{figure*}

\newpage
\section{Detailed Individual-Level Results}
\label{indpredall}
\subsection{Budget Allocation}
\input{tables/sc1_indpred}
\newpage
\subsection{Investment Decision-Making}
\input{tables/sc3_indpred}
\newpage
\subsection{College Jobs}
\input{tables/sc4_indpred}
\newpage
\input{tables/prompt_code}

\end{document}

%% file: tables/new/table_st_sqnsq.tex
\begin{table*}[htpb]

                \caption[]{ Choice Preferences under different Status Quo conditions in the No-Load condition for all three scenarios. SQ = Status Quo, N = Neutral, NSQ = Non-Status Quo.}

                \centering
                \renewcommand{\arraystretch}{1.5} 
                \label{sec:sqb framing vs non-sqb framing}%
                \begin{tabular}{llllllll}%
                    \toprule%
                    \multirow{3}{*}{\textbf{Alternatives}}&\multicolumn{3}{c}{\textbf{Choice rates}}&&\multicolumn{3}{c}{\textbf{Status quo framing vs. non-status quo framing}}\\%
                    \cline{2-4}%
                    \cline{6-8}%
                                                                                                                                    &   &   &   &   &   &   &   \\%
                    &\textbf{SQ}&\textbf{N}&\textbf{NSQ}&& \textbf{\textit{p}}&\textbf{Odds ratio} \newline%
                    \hspace{1pt}\textbf{(95\% CI)}& \textbf{Cohen's h} \newline%
                    \hspace{1pt}\textbf{(95\%CI)}\\%
                    \midrule%
                    \textbf{Scenario 1: Budget allocation ratios} &   &   &   &   &   &   &   \\%
                    60A40H                                                   & 26/51 (0.51) & 14/60 (0.23) & 10/73 (0.14) &  & 0.000 & 0.12 [0.04, 0.34] & 0.79 [0.43, 1.15] \\%
                    50A50H                                                   & 63/73 (0.86) & 46/60 (0.77) & 25/51 (0.49) &  &  &  &  \\%
                    \textbf{Scenario 2: Investment portfolios}    &   &   &   &   &   &   &   \\%
                    Mod. Risk                                                & 43/58 (0.74) & 37/51 (0.73) & 12/54 (0.22) &  & 0.657 & 1.33 [0.38, 4.65] & 0.09 [-0.28, 0.46] \\%
                    High Risk                                                & 42/54 (0.78) & 14/51 (0.27) & 15/58 (0.26) &  &  &  &   \\%
                    \textbf{Scenario 3: College jobs }            &   &   &   &   &   &   &   \\%
                    College A                                                & 49/70 (0.70) & 42/76 (0.55) & 27/51 (0.53) &  & 0.041 & 2.35 [1.03, 5.31] & 0.47 [0.11, 0.83] \\%
                    College B                                                & 24/51 (0.47) & 34/76 (0.45) & 21/70 (0.30) &  &  &  &  \\%
                    \bottomrule%
                \end{tabular}
                \newline
                \label{tab:ST-SQNSQ}
            \end{table*}

%% file: tables/new/table_ct_sqnsq.tex
\begin{table*}[htpb]
                \caption[]{Choice Preferences under different Status Quo conditions in the Load condition for all three scenarios. SQ = Status Quo, N = Neutral, NSQ = Non-Status Quo.}
                \centering
                \renewcommand{\arraystretch}{1.5} 
                \label{sec:sqb framing vs non-sqb framing}%
                \begin{tabular}{llllllll}%
                    \toprule%
                    \multirow{3}{*}{\textbf{Alternatives}}&\multicolumn{3}{c}{\textbf{Choice rates}}&&\multicolumn{3}{c}{\textbf{Status quo framing vs. non-status quo framing}}\\%
                    \cline{2-4}%
                    \cline{6-8}%
                                                                                                                                    &   &   &   &   &   &   &   \\%
                    &\textbf{SQ}&\textbf{N}&\textbf{NSQ}&& \textit{\textbf{p}}&\textbf{Odds ratio} \newline%
                    \hspace{1pt}\textbf{(95\% CI)}&\textbf{Cohen's h} \newline%
                    \hspace{1pt}\textbf{(95\%CI)}\\%
                    \midrule%
                    \textbf{Scenario 1: Budget allocation ratios} &   &   &   &   &   &   &   \\%
                    60A40H                                                   & 28/59 (0.47) & 14/70 (0.20) & 10/64 (0.16) &  & 0.001 & 0.19 [0.07, 0.51] & 0.81 [0.46, 1.16] \\%
                    50A50H                                                   & 54/64 (0.84) & 56/70 (0.80) & 31/59 (0.53) &  &  &  &  \\%
                    \textbf{Scenario 2: Investment portfolios}    &   &   &   &   &   &   &   \\%
                    Mod. Risk                                                & 41/51 (0.80) & 42/57 (0.74) & 12/56 (0.21) &  & 0.207 & 2.31 [0.63, 8.42] & 0.04 [-0.33, 0.42] \\%
                    High Risk                                                & 44/56 (0.79) & 15/57 (0.26) & 10/51 (0.20) &  &  &  &   \\%
                    \textbf{Scenario 3: College jobs }            &   &   &   &   &   &   &   \\%
                    College A                                                & 50/70 (0.71) & 46/80 (0.57) & 28/49 (0.57) &  & 0.001 & 5.20 [1.98, 13.64] & 0.59 [0.22, 0.95] \\%
                    College B                                                & 21/49 (0.43) & 34/80 (0.42) & 20/70 (0.29) &  &  &  &  \\%
                    \bottomrule%
                \end{tabular}
                \newline

                \label{tab:CT-SQNSQ}
            \end{table*}

%% file: tables/h3_h1.tex
\begin{table*}[htpb]
\centering
\caption[]{Human and LLM-agent (GPT4.1) effect sizes (Cohen's h) with 95\% confidence intervals. Significance: * p $<$ 0.05, ** p $<$ 0.01, *** p $<$ 0.001.}
\begin{tabular}{@{}cccccc@{}}
\toprule
\multirow{2}{*}{\textbf{Choice Problem}} & \multirow{2}{*}{\textbf{Prior Dialogue}} & \multirow{2}{*}{\textbf{Human}} & \multicolumn{3}{c}{\textbf{Agent}} \\ \cmidrule(l){4-6} 
 &  &  & \textbf{Human Likeness 1} & \textbf{Human Likeness 2} & \textbf{Human Likeness 3} \\ \midrule
\multirow{2}{*}{\begin{tabular}[c]{@{}c@{}}\textbf{Budget} \\ \textbf{Allocation}\end{tabular}} & Simple Dialogue & 0.779 [0.602, 0.956]*** & 1.503 [1.325, 1.680]*** & 0.990 [0.813, 1.168]*** & 2.781 [2.603, 2.958]*** \\ \cmidrule(l){2-6}
 & Complex Dialogue & 0.794 [0.618, 0.969]*** & 0.891 [0.715, 1.066]*** & 0.596 [0.421, 0.772]** & 2.783 [2.608, 2.959]*** \\ \midrule
\multirow{2}{*}{\textbf{Investment}} & Simple Dialogue & 0.082 [-0.101, 0.266] & 0.147 [-0.036, 0.331] & 0.007 [-0.177, 0.190] & 2.766 [2.583, 2.950]*** \\ \cmidrule(l){2-6}
 & Complex Dialogue & 0.043 [-0.145, 0.231] & 0.009 [-0.179, 0.197] & 0.009 [-0.179, 0.197] & 2.758 [2.570, 2.946]*** \\ \midrule
\multirow{2}{*}{\textbf{College Jobs}} & Simple Dialogue & 0.463 [0.284, 0.642]* & 2.776 [2.597, 2.955]*** & 2.777 [2.598, 2.956]*** & 2.193 [2.014, 2.373]*** \\ \cmidrule(l){2-6}
 & Complex Dialogue & 0.577 [0.396, 0.758]** & 2.773 [2.592, 2.954]*** & 2.773 [2.592, 2.954]*** & 2.625 [2.444, 2.806]*** \\ 
\bottomrule
\end{tabular}%

\label{tab:h3_h1}
\end{table*}

%% file: others/chat_transcript.tex
\begin{framed}
\footnotesize
\label{DialogueTranscripts}

\textbf{\textit{User:}} You are a participant in a research study. The following is your demographic information: \{`Age': 25, `Sex': Female, `Ethnicity simplified': Black, `Country of birth': United Kingdom, `Country of residence': United Kingdom, `Nationality': United Kingdom, `Language': English, `Student status': None, `Employment status': Full-Time\}. Your task is to respond to the user. Study Instructions: The user will ask various questions to understand your preferences on a topic. At some point, the user will change the topic and ask you a different scenario-based question. Please pay attention and ensure you understand the question before answering. Later in this study, you must answer three extra questions to identify if you understood the scenario-based question and all the options correctly. After the scenario-based question, the user will present various attributes. You need to compare and select the best. When the user asks you to remember specific details, please memorise them carefully, as you will need to recall them later in the study. Shall we begin? Please respond with 'yes' to start the conversation.  

\textbf{\textit{Assistant:}}  Yes, let's begin!  

\textbf{\textit{User:}}  Hello! I'm here to understand your preferences through various Scenarios. Shall we start?  

\textbf{\textit{Assistant:}}  Yes, lets begin  

\textbf{\textit{User:}}  Do you like listening to pop music?  

\textbf{\textit{Assistant:}}  Yes i do  

\quad \vdots 









\textbf{\textit{User:}}  Do you prefer music from the 90s?  

\textbf{\textit{Assistant:}}  A little  

\textbf{\textit{User:}}  Thanks for sharing! Now, let's get started with some specific scenarios to understand your preferences.  
The first artist performs three live shows, is paid 2000 units per show, and has a 4-star rating. The second artist performs twice as many shows, with the same pay and rating. Which artist is better, and why?  

\textbf{\textit{Assistant:}}  The second artist  

\quad \vdots 





\textbf{\textit{User:}}  Remember number of live performances, remuneration, and the star rating of the fourth artist. Now, let’s move on to a different scenario. You are currently an assistant professor at College A in the east coast. Recently, you have been approached by a colleague at another university with a job opportunity. When evaluating teaching job offers, people typically consider the salary, the reputation of the school, the location of the school, and the likelihood of getting tenure (tenure is a permanent job contract that can only be terminated for cause or under extraordinary circumstances). Your choices are:  

- Remain at College A: east coast, very prestigious school, high salary, fair chance of tenure.  

- Move to College B: west coast, low prestige school, high salary, good chance of tenure.  

\textbf{\textit{Assistant:}} $\cdots$ 

\normalsize
\end{framed}

%% file: tables/ind_pred.tex
\begin{table}[htpb]
\caption[]{This table shows the precision of prediction of human decision by different LLMs at the individual level for each Human-Likeness level.
}
\label{tab:individual-hl}
\centering
\begin{tabular}{lccc}
\toprule
\textbf{Model} & \textbf{HL1} & \textbf{HL2} & \textbf{HL3} \\ 
\midrule
\textbf{gpt4.1} & 0.675 & 0.685 & 0.553 \\
\textbf{gpt4.1-mini} & 0.635 & 0.655 & 0.553 \\
\textbf{gpt5} & 0.532 & 0.595 & 0.615 \\
\textbf{gpt5-mini} & 0.456 & 0.484 & 0.545 \\ 
\bottomrule
\end{tabular}
\end{table}

%% file: tables/precision.tex

\begin{table*}[htpb]
\centering
\caption[]{Confusion matrices for HL1, HL2, HL3 (Biased/Not Biased).}
\begin{tabular}{ccc}
\begin{tabular}{lcc} \toprule\multicolumn{1}{c}{} & \multicolumn{2}{c}{\textbf{LLM-Agent (HL1)}} \\ \cmidrule(lr){2-3} \textbf{Human} & \textbf{Not Biased} & \textbf{Biased} \\\midrule \textbf{Not Biased} & 2 & 0 \\ \textbf{Biased} & 0 & 4 \\ \cmidrule(lr){1-3} \multicolumn{2}{r}{\textbf{Precision}} & 1.00 \\ \bottomrule\end{tabular} & \begin{tabular}{lcc} \toprule\multicolumn{1}{c}{} & \multicolumn{2}{c}{\textbf{LLM-Agent (HL2)}} \\ \cmidrule(lr){2-3} \textbf{Human} & \textbf{Not Biased} & \textbf{Biased} \\\midrule \textbf{Not Biased} & 2 & 0 \\ \textbf{Biased} & 0 & 4 \\ \cmidrule(lr){1-3} \multicolumn{2}{r}{\textbf{Precision}} & 1.00 \\ \bottomrule\end{tabular} & \begin{tabular}{lc} \toprule\multicolumn{1}{c}{} & \multicolumn{1}{c}{\textbf{LLM-Agent (HL3)}} \\ \cmidrule(lr){2-2} \textbf{Human} & \textbf{Biased} \\\midrule \textbf{Not Biased} & 2 \\ \textbf{Biased} & 4 \\ \cmidrule(lr){1-2} \multicolumn{1}{r}{\textbf{Precision}} & 0.67 \\ \bottomrule\end{tabular} \\
\end{tabular}

\label{tab:confusion_matrices}
\end{table*}

%% file: tables/h3_h2.tex
\begin{table*}[htpb]
\caption[]{Comparison of z-values and confidence intervals for effect size differences (Simple vs Complex Dialogue). Significance: * $p < 0.05$, ** $p < 0.01$, *** $p < 0.001$.}
\centering
\begin{tabular}{lcccc}
\toprule
\textbf{Choice Problem} & \textbf{Human} & \textbf{HL1} & \textbf{HL2} & \textbf{HL3} \\
\midrule
\textbf{Budget Allocation }& 0.12 [-0.23, 0.26] & -4.81 [-0.86, -0.36]*** & -3.09 [-0.64, -0.14]** & 0.02 [-0.25, 0.25] \\
\textbf{Investment} & -0.29 [-0.30, 0.22] & -1.03 [-0.40, 0.12] & 0.01 [-0.26, 0.26] & -0.06 [-0.27, 0.25] \\
\textbf{College Jobs} & 0.88 [-0.14, 0.37] & -0.02 [-0.26, 0.25] & -0.03 [-0.26, 0.25] & 3.32 [0.18, 0.69]** \\
\bottomrule
\end{tabular}

\label{tab:h3_h2}
\end{table*}

%% file: tables/precision_and_mad.tex
\begin{table}[htpb]
\centering
\caption[]{Precision and Mean Absolute Difference (MD) for each model.}
\begin{tabular}{lccc|ccc}
\toprule
\multirow{2}{*}{\textbf{Model}} & \multicolumn{3}{c|}{\textbf{Precision}} & \multicolumn{3}{c}{\textbf{MD}} \\ \cmidrule(lr){2-4} \cmidrule(lr){5-7}
 & \textbf{HL1} & \textbf{HL2} & \textbf{HL3} & \textbf{HL1} & \textbf{HL2} & \textbf{HL3} \\ \midrule
\textbf{gpt4.1-mini} & 1.000 & 1.000 & 0.667 & 2.793 & 0.883 & 1.023 \\
\textbf{gpt4.1} & 1.000 & 1.000 & 0.667 & 2.190 & 1.473 & 0.923 \\
\textbf{gpt5-mini} & 1.000 & 1.000 & 0.667 & 2.863 & 2.430 & 0.943 \\
\textbf{gpt5} & 0.000 & 0.500 & 0.667 & 1.223 & 1.263 & 0.673 \\
\bottomrule
\end{tabular}%

\label{tab:precision_mad_all_models}
\end{table}

%% file: tables/ablation.tex
\begin{table*}[htpb]
\centering
\caption[]{Ablation study results showing precision, mean absolute deviation (MD), and root mean squared error (RMSE) when isolating or removing prior dialogue components (choice problem only, memory, arithmetic). Lower MD and RMSE indicate closer alignment to observed human behavior in complex prior dialogue conditions, while precision captures biased behavior.}
\begin{tabular}{@{}llcccc@{}}
\toprule
\textbf{Condition} & \textbf{Model} & \textbf{Precision} & \textbf{MD} & \textbf{RMSE} \\ \midrule
\multirow{2}{*}{\textbf{Choice Problem Only}} 
 & GPT-4.1 & 0.800 & 4.957 & 6.939 \\
 & GPT-4.1-mini & 0.500 & 1.037 & 1.057 \\ \midrule
\multirow{2}{*}{\textbf{\begin{tabular}[c]{@{}l@{}}Memory Component \\ Followed by Choice Problem\end{tabular}}} 
 & GPT-4.1 & 0.800 & 5.800 & 5.932 \\
 & GPT-4.1-mini & 1.000 & 6.183 & 8.803 \\ \midrule
\multirow{2}{*}{\textbf{\begin{tabular}[c]{@{}l@{}}Arithmetic Component Only \\ Followed by Choice Problem\end{tabular}}} 
 & GPT-4.1 & 1.000 & 1.100 & 1.104 \\
 & GPT-4.1-mini & 1.000 & 4.137 & 4.577 \\ \bottomrule
\end{tabular}

\label{tab:ablation_table_new}
\end{table*}

%% file: tables/supp_material/all_decision_scenarios.tex
\onecolumn
\section{Decision Scenarios Adapted From Samuelson and Zeckhauser}
\label{sec:appendix-ds}
\subsection{Budget Allocation (SC1)}
\label{sec:appendix-ds-ba}
\subsubsection{Neutral Condition}
\label{sec:appendix-ds-ba-neut1}

\begin{quote}
The National Highway Safety Commission is deciding how to allocate its budget between two safety research programs:
\begin{itemize}
    \item Improving automobile safety (bumpers, body, gas tank configuration, seat-belts), and
    \item Improving the safety of interstate highways (guard rails, grading, highway interchanges, and implementing selective reduced speed limits).
\end{itemize}

Since there is a ceiling on its total spending, it must choose between the options provided below. If you had to make this choice, which of the following will you choose?
 \begin{itemize}
     \item Allocate 60\% to auto safety and 40\% to highway safety
     \item Allocate 50\% to auto safety and 50\% to highway safety
 \end{itemize}
\end{quote}

\subsubsection{Neutral Condition - Alternative Swapped}
\label{sec:appendix-ds-ba-neut2}

\begin{quote}

The National Highway Safety Commission is deciding how to allocate its budget between two safety research programs:
\begin{itemize}
    \item Improving automobile safety (bumpers, body, gas tank configuration, seat-belts), and
    \item Improving the safety of interstate highways (guard rails, grading, highway interchanges, and implementing selective reduced speed limits).
\end{itemize}

Since there is a ceiling on its total spending, it must choose between the options provided below. If you had to make this choice, which of the following will you choose?
 \begin{itemize}
     \item Allocate 50\% to auto safety and 50\% to highway safety.
     \item Allocate 60\% to auto safety and 40\% to highway safety.
 \end{itemize}

\end{quote}

\subsubsection{Status Quo A - 60A40H }
\label{sec:appendix-ds-ba-sq60a40h}

\begin{quote}

The National Highway Safety Commission is deciding how to allocate its budget between two safety research programs:
\begin{itemize}
    \item Improving automobile safety (bumpers, body, gas tank configuration, seat-belts)
    \item Improving the safety of interstate highways (guard rails, grading, highway interchanges, and implementing selective reduced speed limits).
\end{itemize}

Currently, the commission allocates approximately 60\% of its funds to auto safety and 40\% of its funds to highway safety.
Since there is a ceiling on its total spending, it must choose between the options provided below. If you had to make this choice, which of the following will you choose?
 \begin{itemize}
     \item Maintain present budget amounts for the programs.
     \item Decrease auto program by 10\% and raise highway program by like amount.
 \end{itemize}
    
\end{quote}

\subsubsection{Status Quo B - 50A50H }
\label{sec:appendix-ds-ba-sq50a50h}

\begin{quote}

The National Highway Safety Commission is deciding how to allocate its budget between two safety research programs: 
\begin{itemize}
    \item Improving automobile safety (bumpers, body, gas tank configuration, seat-belts)
    \item Improving the safety of interstate highways (guard rails, grading, highway interchanges, and implementing selective reduced speed limits).
\end{itemize}

Currently, the commission allocates approximately 50\% of its funds to auto safety and 50\% of its funds to highway safety.
Since there is a ceiling on its total spending, it must choose between the options provided below. If you had to make this choice, which of the following will you choose? 

 \begin{itemize}
     \item Maintain present budget amounts for the programs.
     \item Increase auto program by 10\% and lower highway program by like amount.
 \end{itemize}

\end{quote}

\subsection{Investment Decision Making (SC3)}
\label{sec:appendix-ds-idm}

\subsubsection{Neutral Condition}
\label{sec:appendix-ds-idm-neut1}

\begin{quote}

You are a serious reader of the financial pages but until recently have had few funds to invest. That is when you inherited a large sum of money from your great uncle. You are considering different portfolios.
Your choices are:
 \begin{itemize}
     \item Invest in moderate-risk Company A. Over a year's time, the stock has .5 chance of increasing 30\% in value, a .2 chance of being unchanged, and a .3 chance of declining 20\% in value.
     \item Invest in high-risk Company B. Over a year's time, the stock has a .4 chance of doubling in value, a .3 chance of being unchanged, and a .3 chance of declining 40\% in value.
 \end{itemize}
    
\end{quote}

\subsubsection{Neutral Condition - Alternative Swapped}
\label{sec:appendix-ds-idm-neut2}

\begin{quote}

You are a serious reader of the financial pages but until recently have had few funds to invest. That is when you inherited a large sum of money from your great uncle. You are considering different portfolios.
Your choices are:
 \begin{itemize}
     \item Invest in high-risk Company B. Over a year's time, the stock has a .4 chance of doubling in value, a .3 chance of being unchanged, and a .3 chance of declining 40\% in value.
     \item Invest in moderate-risk Company A. Over a year's time, the stock has .5 chance of increasing 30\% in value, a .2 chance of being unchanged, and a .3 chance of declining 20\% in value.
 \end{itemize}

\end{quote}

\subsubsection{Status Quo A - Moderate Risk}
\label{sec:appendix-ds-idm-modrisk}

\begin{quote}
    You are a serious reader of the financial pages but until recently have had few funds to invest. That is when you inherited a portfolio of cash and securities from your great uncle. A significant portion of this portfolio is invested in moderate-risk Company A. You are deliberating whether to leave the portfolio intact or change it by investing in other securities. (The tax and broker commission consequences of any change are insignificant.)
Your choices are:
 \begin{itemize}
     \item Retain the investment in moderate-risk Company A. Over a year's time, the stock has .5 chance of increasing 30\% in value, a .2 chance of being unchanged, and a .3 chance of declining 20\% in value.
     \item Invest in high-risk Company B. Over a year's time, the stock has a .4 chance of doubling in value, a .3 chance of being unchanged, and a .3 chance of declining 40\% in value.
 \end{itemize}
\end{quote}

\subsubsection{Status Quo B - High Risk}
\label{sec:appendix-ds-idm-highrisk}

\begin{quote}
    You are a serious reader of the financial pages but until recently have had few funds to invest. That is when you inherited a portfolio of cash and securities from your great uncle. A significant portion of this portfolio is invested in high-risk Company B. You are deliberating whether to leave the portfolio intact or change it by investing in other securities. (The tax and broker commission consequences of any change are insignificant.)
Your choices are:
 \begin{itemize}
     \item Retain the investment in high-risk Company B. Over a year's time, the stock has a .4 chance of doubling in value, a .3 chance of being unchanged, and a .3 chance of declining 40\% in value.
     \item Invest in moderate-risk Company A. Over a year's time, the stock has a .5 chance of increasing 30\% in value, a .2 chance of being unchanged, and a .3 chance of declining 20\% in value.
 \end{itemize}
\end{quote}

\subsection{College Jobs (SC4)}
\label{sec:appendix-ds-cj}

\subsubsection{Neutral Condition}
\label{sec:appendix-ds-cj-neut1}

\begin{quote}
    Having just completed your graduate degree, you have two offers of teaching jobs in hand.
When evaluating teaching job offers, people typically consider the salary, the reputation of the school, the location of the school, and the likelihood of getting tenure (tenure is permanent job contract that can only be terminated for cause or under extraordinary circumstances).
Your choices are:
 \begin{itemize}
     \item College A: east coast, very prestigious school, high salary, fair chance of tenure.
     \item College B: west coast, low prestige school, high salary, good chance of tenure.
 \end{itemize}
\end{quote}

\subsubsection{Neutral Condition - Alternative Swapped}
\label{sec:appendix-ds-cj-neut2}

\begin{quote}
    Having just completed your graduate degree, you have two offers of teaching jobs in hand.
When evaluating teaching job offers, people typically consider the salary, the reputation of the school, the location of the school, and the likelihood of getting tenure (tenure is permanent job contract that can only be terminated for cause or under extraordinary circumstances).
Your choices are:
 \begin{itemize}
     \item College B: west coast, low prestige school, high salary, good chance of tenure.
     \item College A: east coast, very prestigious school, high salary, fair chance of tenure.
 \end{itemize}
\end{quote}

\subsubsection{Status Quo A - College A}
\label{sec:appendix-ds-cj-colla}
\begin{quote}
You are currently an assistant professor at College A in the east coast. Recently, you have been approached by colleague at other university with job opportunity.
When evaluating teaching job offers, people typically consider the salary, the reputation of the school, the location of the school, and the likelihood of getting tenure (tenure is permanent job contract that can only be terminated for cause or under extraordinary circumstances).
Your choices are:
 \begin{itemize}
     \item Remain at College A: east coast, very prestigious school, high salary, fair chance of tenure.
     \item Move to College B: west coast, low prestige school, high salary, good chance of tenure.
 \end{itemize}
 \end{quote}
\begin{quote}
    
\end{quote}

\subsubsection{Status Quo B - College B}
\label{sec:appendix-ds-cj-collb}
\begin{quote}
    
You are currently an assistant professor at College B in the west coast. Recently, you have been approached by colleague at other university with job opportunity.
When evaluating teaching job offers, people typically consider the salary, the reputation of the school, the location of the school, and the likelihood of getting tenure (tenure is permanent job contract that can only be terminated for cause or under extraordinary circumstances). 
Your choices are:
 \begin{itemize}
     \item Remain at College B: west coast, low prestige school, high salary, good chance of tenure.
     \item Move to College A: east coast, very prestigious school, high salary, fair chance of tenure.
 \end{itemize}
\end{quote}

%% file: others/text_adj_sqb.tex
\subsection{Textual Adjustments to Original Decision Scenarios}
\label{appendix:decision_scenario-manipulations}

We made a few syntactic adjustments to the decision scenarios to ensure consistency across different experimental conditions. Below, we outline the key modifications made to the phrasing, structure, and presentation of the scenarios.

In the neutral condition of the college job scenarios, we reduced the number of teaching job offers from four to two.
Similarly, in the status quo condition of the college jobs scenario, we changed the text ``\textit{colleagues at other universities with job opportunities}'' to ``\textit{colleague at other university with job opportunity}''. In all the decision scenarios, we moved from an ordered list presentation to bullet points. 

The replication study by ~\citet{xiao_revisiting_2021} included a method to assess whether participants understood the decision scenario. Participants were first shown the scenario and asked various related questions before being presented with the decision alternatives.
To maintain conversational fluidity, we asked the comprehension questions later in the survey (as \textit{Decision Scenario Attention Check}), and not during the chatbots' interaction. 
Accuracy on the Decision Scenario Attention Check was used as a filtering criterion. 
Additionally, we asked participants whether they had encountered the decision scenario before, recorded as \textit{Decision Scenario Familiarity}. 
To mitigate potential learning effects, this variable was also used to exclude data from participants with prior exposure.

%% file: tables/supp_material/simple_task_questions.tex
\section{Prior Dialogue}
\label{sec:appendix-prior_dialogue}
\subsection{Simple Dialogue}
\label{sec:appendix-simple_task-prior_dialogue}
\begin{table*}[htpb]
\caption[]{Yes/No questions of preference elicitation tasks, designed using a less conservative dialogue strategy, across five different domains of the SGD Dataset. Please enter `` I don't know'' as an attention check. }
\centering
\begin{tabular}{lll}
\toprule
\textbf{Domain}                    & \textbf{Attribute}           & \textbf{Yes/No Questions}                                                          \\ \midrule
\multirow{6}{*}{\textbf{Real Estate}}    & Budget                       & Do you have a specific budget for the home?                                        \\
                                   & Location                     & Are you looking for a home in a specific location?                                 \\
                                   & Number of Bedrooms           & Do you need more than 3 bedrooms?                                                  \\
                                   & Number of Bathrooms          & Is having 2 or more bathrooms important to you?                                    \\
 &
  Type of Home (Apartment, House, etc.) &
  \begin{tabular}[c]{@{}l@{}}Are you looking specifically for a detached house? \\ Please enter “I don't know” only.\end{tabular} \\
                                   & Size (Square Footage)        & Do you prefer homes larger than 2000 square feet?                                  \\ \hline
\multirow{6}{*}{\textbf{Music}}    & Genre Preference             & Do you like listening to pop music?                                                \\
                                   & Language of Lyrics           & Do you prefer music with lyrics in English?                                        \\
                                   & Live Performances            & Are you interested in live music performances?                                     \\
                                   & Instruments Focused          & Do you enjoy instrumental music?                                                   \\
                                   & Artist-Specific              & Do you like music from specific artists? Please enter “I don't know” only.         \\
                                   & Era (e.g., 80s, 90s)         & Do you prefer music from the 90s?                                                  \\ \hline
\multirow{6}{*}{\textbf{Movies}}   & Genre Preference             & Do you like action movies?                                                         \\
                                   & Language Preference          & Do you prefer movies in English?                                                   \\
                                   & Director/Actor Specific      & Are you a fan of movies by a specific director or actor?                           \\
                                   & Streaming Service Preference & Do you prefer to watch movies on Netflix?                                          \\
                                   & High Definition (HD)         & Do you prefer movies in high definition? Please enter “I don't know” only.         \\
                                   & Subtitles                    & Do you need subtitles when watching movies?                                        \\ \hline
\multirow{6}{*}{\textbf{Calendar}} & Daily Reminders              & Do you need daily reminders?                                                       \\
                                   & Meeting Notifications        & Do you want notifications for upcoming meetings?                                   \\
                                   & Recurring Events             & Do you often schedule recurring events?                                            \\
                                   & Shared Calendars             & Do you use shared calendars with others?                                           \\
 &
  Event Location &
  \begin{tabular}[c]{@{}l@{}}Is specifying the location of events important to you? \\ Please enter “I don't know” only.\end{tabular} \\
                                   & Time Zone Adjustments        & Do you need automatic time zone adjustments for events?                            \\ \hline
\multirow{6}{*}{\textbf{Banks}}    & Online Banking               & Do you prefer to use online banking?                                               \\
                                   & Mobile Banking               & Do you want to manage your accounts through mobile banking apps?                   \\
                                   & Bank Branch Proximity        & Is it important that a bank branch is close to your home or work?                  \\
                                   & ATM Availability             & Is having nearby ATMs a priority for you?                                          \\
                                   & Interest Rates               & Are competitive interest rates important to you? Please enter “I don't know” only. \\
                                   & Fee-Free Accounts            & Do you prefer accounts with no fees?                                               \\ \hline
\multirow{6}{*}{\textbf{Apps}} &
  Security &
  Do you prefer end-to-end encryption (security) for enhanced privacy? \\
                                   & Cross-Platform               & Is cross-platform compatibility (iOS, Android, desktop) important to you?          \\
 &
  File Sharing &
  \begin{tabular}[c]{@{}l@{}}Do you value the ability to send multimedia \\ files like photos, videos, and documents?\end{tabular} \\
 &
  Group Messaging &
  \begin{tabular}[c]{@{}l@{}}Is group messaging and the ability to create large groups \\ important for your communication needs?\end{tabular} \\
 &
  Calling Features &
  \begin{tabular}[c]{@{}l@{}}Do you prefer apps with voice and video calling features? \\ Please enter “I don't know” only.\end{tabular} \\
                                   & Customisability              & Are customisable notifications and settings important to you?                      \\ 
\bottomrule
\end{tabular}%
\label{tab:simple_task_questions}
\end{table*}

%% file: tables/supp_material/complex_task_questions.tex
\onecolumn
\subsection{Complex Dialogue}
\label{sec:appendix-complex_task-prior_dialogue}

\begin{table}[htpb]
\centering
\caption[]{Complex dialogue for Home Property domain}
\label{tab:appendix-complex-task}
\resizebox{\textwidth}{!}{%
\begin{tabular}{@{}ccccc@{}}
\toprule
\multirow{2}{*}{\textbf{\begin{tabular}[c]{@{}c@{}}Domain\\ (Home \\ Property)\end{tabular}}} & \multicolumn{3}{c}{\textbf{Attribute}} & \multirow{2}{*}{\textbf{Questions}} \\ \cmidrule(lr){2-4}
 & \textbf{\begin{tabular}[c]{@{}c@{}}Number of \\ Bedrooms\end{tabular}} & \textbf{\begin{tabular}[c]{@{}c@{}}Size \\ (Sq ft.)\end{tabular}} & \textbf{Property Reviews} &  \\ \midrule
1st & Three Rooms & 2000 & Four star & \begin{tabular}[c]{@{}c@{}}In the following scenario choose \\ from various property recommendations.\\ The first property has three bedrooms, \\ 2000 square feet, and a 4-star rating.\end{tabular} \\ \midrule
2nd & Two times 1st & Same as first & Same as first & \begin{tabular}[c]{@{}c@{}}The second property has twice the number of\\ bedrooms and with the same size and rating. \\ Which one do you prefer, and why?\end{tabular} \\ \midrule
3rd & Same as the second & Half of first & Same as first & \begin{tabular}[c]{@{}c@{}}The third property has the same \\ number of bedrooms as the second one \\ but is half the size of the first one,\\ with the same rating as the first. \\ Which one do you prefer, and why?\end{tabular} \\ \midrule
4th & Same as the second & Same as third & \begin{tabular}[c]{@{}c@{}}One star less \\ than the first\end{tabular} & \begin{tabular}[c]{@{}c@{}}The fourth property has the same \\ number of bedrooms as the second, \\ the same size as the third, but \\ one less star rating than the first. \\ Which one do you prefer, and why?\\ Remember the details \\ of the fourth property. \\ Specific information \\ will be requested later.\end{tabular} \\ \bottomrule
\end{tabular}%
}
\end{table}

\newpage

\begin{table}[htpb]
\centering
\caption[]{Complex dialogue for the Music Artist domain}
\label{tab:appendix-complex-task}
\resizebox{\textwidth}{!}{%
\begin{tabular}{@{}ccccc@{}}
\toprule
\multirow{2}{*}{\textbf{\begin{tabular}[c]{@{}c@{}}Domain\\ (Music\\ Artist)\end{tabular}}} & \multicolumn{3}{c}{\textbf{Attribute}} & \multirow{2}{*}{\textbf{Questions}} \\ \cmidrule(lr){2-4}
 & \textbf{\begin{tabular}[c]{@{}c@{}}Live \\ Performances\end{tabular}} & \textbf{\begin{tabular}[c]{@{}c@{}}Artist \\ Remuneration \\ for Show\end{tabular}} & \textbf{Artist-Specific} &  \\ \midrule
1st & Three & 2000 Units & Four star & \begin{tabular}[c]{@{}c@{}}In the following scenario choose\\ from various artist recommendations.\\ The first artist performs\\ three live shows, \\ is paid 2000 units per show, \\ and has a 4-star rating.\end{tabular} \\ \midrule
2nd & 2 times 1st & Same as first & Same as first & \begin{tabular}[c]{@{}c@{}}The second artist performs \\ twice as many shows, \\ with the same pay and rating. \\ Which artist do you prefer, and why?\end{tabular} \\ \midrule
3rd & Same as the second & Half of first & Same as first & \begin{tabular}[c]{@{}c@{}}The third artist performs the same\\ number of shows as the second, \\ earns half the pay of the first artist, \\ but has the same rating as the first. \\ Which artist do you prefer, and why?\end{tabular} \\ \midrule
4th & Same as the second & Same as third & \begin{tabular}[c]{@{}c@{}}One star less \\ than the first\end{tabular} & \begin{tabular}[c]{@{}c@{}}The fourth artist performs the \\ same number of shows as the second, \\ earns the same pay as the third, but \\ has two stars less than the first artist. \\ Which artist do you prefer, and why? Remember the \\ details of the fourth artist. \\ Specific information\\ will be requested later.\end{tabular} \\ \bottomrule
\end{tabular}%
}
\end{table}

\newpage

\begin{table}[htpb]
\centering
\caption[]{Complex dialogue for the Movies - Streaming Service domain}
\label{tab:appendix-complex-task}
\resizebox{\textwidth}{!}{%
\begin{tabular}{@{}ccccc@{}}
\toprule
\multirow{2}{*}{\textbf{\begin{tabular}[c]{@{}c@{}}Domain\\ (Movies - \\ Streaming  Service)\end{tabular}}} & \multicolumn{3}{c}{\textbf{Attribute}} & \multirow{2}{*}{\textbf{Questions}} \\ \cmidrule(lr){2-4}
 & \textbf{\begin{tabular}[c]{@{}c@{}}Number of \\ Parallel Devices\end{tabular}} & \textbf{Library Size} & \textbf{Service Rating} &  \\ \midrule
1st & Three & 2000 Movies & Four star & \begin{tabular}[c]{@{}c@{}}In the following scenario choose from \\ various streaming service recommendations.\\ The first streaming service supports \\ 3 parallel devices, \\ has a library of 2000 movies,\\ and is rated 4 stars.\end{tabular} \\ \midrule
2nd & 2 times 1st & Same as first & Same as first & \begin{tabular}[c]{@{}c@{}}The second service supports \\ twice as many devices, \\ with the same library size and rating. \\ Which service do you prefer, and why?\end{tabular} \\ \midrule
3rd & Same as the second & Half of first & Same as first & \begin{tabular}[c]{@{}c@{}}The third streaming service supports the\\ same number of devices as the second, has \\ half the library size of the first,\\ but has the same rating as the first. \\ Which service do you prefer, and why?\end{tabular} \\ \midrule
4th & Same as the second & Same as third & \begin{tabular}[c]{@{}c@{}}One star less\\ than the first\end{tabular} & \begin{tabular}[c]{@{}c@{}}The fourth streaming service supports\\ the same number of devices as the second, \\ has the same library size as the third,\\ but is rated one star less than the first. \\ Which service do you prefer, and why? \\ Remember the details of the fourth service. \\ Specific information will be requested later.\end{tabular} \\ \bottomrule
\end{tabular}%
}
\end{table}

\newpage

\begin{table}[htpb]
\centering
\caption[]{Complex dialogue for the Calendar App domain}
\label{tab:appendix-complex-task}
\resizebox{\textwidth}{!}{%
\begin{tabular}{@{}ccccc@{}}
\toprule
\multirow{2}{*}{\textbf{\begin{tabular}[c]{@{}c@{}}Domain\\ (Calendar\\ App)\end{tabular}}} & \multicolumn{3}{c}{\textbf{Attribute}} & \multirow{2}{*}{\textbf{Questions}} \\ \cmidrule(lr){2-4}
 & \textbf{\begin{tabular}[c]{@{}c@{}}Calendar Syncing \\ Across Devices\end{tabular}} & \textbf{\begin{tabular}[c]{@{}c@{}}Managed \\ Tasks \\ Per Year\end{tabular}} & \textbf{\begin{tabular}[c]{@{}c@{}}Event Privacy \\ Rating\end{tabular}} &  \\ \midrule
1st & 3 & 2000 & Four star & \begin{tabular}[c]{@{}c@{}}In the following scenario choose from \\ Various Apps recommendations for calendar.\\ The first calendar app can sync\\ across three devices, \\ manages 2000 tasks per year,\\ and has a 4-star privacy rating.\end{tabular} \\ \midrule
2nd & 2 times 1st & Same as first & Same as first & \begin{tabular}[c]{@{}c@{}}The second app syncs across two devices, \\ manages the same number of tasks,\\ and has the same privacy rating. \\ Which app do you prefer, and why?\end{tabular} \\ \midrule
3rd & Same as Second & Half of first & Same as first & \begin{tabular}[c]{@{}c@{}}The third app syncs across the same\\ number of devices as the second app, \\ but manages half as\\ many tasks as the first app, \\ with the same privacy rating as the first. \\ Which app do you prefer, and why?\end{tabular} \\ \midrule
4th & Same as Second & Same as third & \begin{tabular}[c]{@{}c@{}}One star less\\ than the first\end{tabular} & \begin{tabular}[c]{@{}c@{}}The fourth app syncs across the same number \\ of devices as the second, manages the same\\ number of tasks as the third, but has one less \\ star in privacy rating compared to the first. \\ Which app do you prefer, and why?\\ Remember the details of the fourth App. \\ Specific information will be requested later.\end{tabular} \\ \bottomrule
\end{tabular}%
}
\end{table}

\newpage

\begin{table}[htpb]
\centering
\caption[]{Complex dialogue for the Bank domain}
\label{tab:appendix-complex-task}
\resizebox{\textwidth}{!}{%
\begin{tabular}{@{}ccccc@{}}
\toprule
\multirow{2}{*}{\textbf{\begin{tabular}[c]{@{}c@{}}Domain\\ (Bank)\end{tabular}}} & \multicolumn{3}{c}{\textbf{Attribute}} & \multirow{2}{*}{\textbf{Questions}} \\ \cmidrule(lr){2-4}
 & \textbf{\begin{tabular}[c]{@{}c@{}}Bank Branch \\ Proximity\end{tabular}} & \textbf{Interest Rates} & \textbf{\begin{tabular}[c]{@{}c@{}}Fee-Free \\ Accounts\\ Rating\end{tabular}} &  \\ \midrule
1st & 3km & 2000 units & Four & \begin{tabular}[c]{@{}c@{}}In the following scenario choose \\ from various banks recommendations.\\ The first bank is 3 km away,\\ offers 2000 units of interest, \\ and has a four-star rating for fee-free accounts.\end{tabular} \\ \midrule
2nd & 2 times 1st & Same as first & Same as first & \begin{tabular}[c]{@{}c@{}}The second bank is twice as far away, \\ offers the same amount of interest,\\ and has the same fee-free account rating. \\ Which bank would you prefer, and why?\end{tabular} \\ \midrule
3rd & Same as the second & Half of first & Same as first & \begin{tabular}[c]{@{}c@{}}The third bank is as far away as the second bank, \\ offers half the amount of interest as \\ the first bank, but has the same fee-free\\ account rating as the first bank. \\ Which bank would you prefer, and why?\end{tabular} \\ \midrule
4th & Same as the second & Same as third & \begin{tabular}[c]{@{}c@{}}One star less \\ than the first\end{tabular} & \begin{tabular}[c]{@{}c@{}}The fourth bank is as far\\ away as the second bank, \\ offers the same amount of interest as the third bank, \\ but has one star less in fee-free account rating\\ compared to the first bank. \\ Which bank would you prefer, and why?\\ Remember the details of the fourth Bank. \\ Specific information will be requested later.\end{tabular} \\ \bottomrule
\end{tabular}%
}
\end{table}

\newpage

\begin{table}[htpb]
\centering
\caption[]{Complex dialogue for the Messaging App domain}
\label{tab:appendix-complex-task}
\resizebox{\textwidth}{!}{%
\begin{tabular}{@{}ccccc@{}}
\toprule
\multirow{2}{*}{\textbf{\begin{tabular}[c]{@{}c@{}}Domain\\ (Messaging App)\end{tabular}}} & \multicolumn{3}{c}{\textbf{Attribute}} & \multirow{2}{*}{\textbf{Questions}} \\ \cmidrule(lr){2-4}
 & \textbf{\begin{tabular}[c]{@{}c@{}}Number of \\ Simultaneous \\ Devices\end{tabular}} & \textbf{\begin{tabular}[c]{@{}c@{}}Messages \\ Per Day\end{tabular}} & \textbf{\begin{tabular}[c]{@{}c@{}}Security \\ Rating\end{tabular}} &  \\ \midrule
1st & 3 & 2000 & Four star & \begin{tabular}[c]{@{}c@{}}In the following scenario, choose from \\ various messaging app recommendations.\\ The first messaging app allows access \\ on three devices, supports 2000 messages per day, \\ and has a 4-star security rating.\end{tabular} \\ \midrule
2nd & 2 times 1st & Same as first & Same as first & \begin{tabular}[c]{@{}c@{}}The second app allows access on\\ twice as many devices, \\ supports the same number of messages, \\ and has the same security rating. \\ Which app do you prefer, and why?\end{tabular} \\ \midrule
3rd & Same as the second & Half of first & Same as first & \begin{tabular}[c]{@{}c@{}}The third app allows access on the same \\ number of devices as the second app, \\ but supports half as many messages as the first app, \\ with the same security rating as the first. \\ Which app do you prefer, and why?\end{tabular} \\ \midrule
4th & Same as the second & Same as third & One star less than the first & \begin{tabular}[c]{@{}c@{}}The fourth app allows access on the \\ same number of devices as the second, \\ supports the same number \\ of messages as the third, \\ but has one less star in \\ security rating compared to the first. \\ Which app do you prefer, and why?\\ Remember the details\\ of the fourth app\\ specific information \\ will be requested later.\end{tabular} \\ \midrule
\multicolumn{1}{l}{} &  &  &  & 
\end{tabular}%
}
\end{table}

\newpage

%% file: others/Additional_DOX.tex
\section{Additional Experiment Design Elements}
\label{appendix:full_dox}
\subsection{Recall Task performance}
Participants completed a recall task after the dialogue to assess their performance on the memory component. The task required participants to recall the answers for the computation they performed over the arithmetic and memory tasks.  This measure allowed us to quantify the effectiveness of different complex dialogues. Recall accuracy was scored and compared across experimental conditions to understand how dialogue structure may impact decision outcomes.
\subsection{Decision Scenarios Recall Task}
A dedicated recall task specifically targeted information presented within the decision scenarios embedded in the dialogues. Drawing inspiration from established replication studies ~\citep{xiao_revisiting_2021}, we adapted previously validated scenario descriptions to our conversational context. Participants’ recall of scenario-specific details—such as available options, consequences, and stated preferences—was measured to examine their comprehension and retention. Since we cannot expose the participants to the decision scenarios before the dialogue and ask them if they understood the question. We implemented the validation as a recall task.
\subsection{Domain Familiarity}
We assessed each participant’s familiarity with the domain of the dialogues to control for prior knowledge effects. Moreover, less familiar domains in the complex task can result in additional cognitive load. Before commencing the main tasks, participants rated their level of experience or comfort with the relevant subject matter using a 7-point Likert scale. These self-reports allowed us to investigate whether domain familiarity influenced recall performance and to include it as a covariant in subsequent analyses. 
\subsection{External Tool Usage}
To control for potential confounds from outside assistance, participants were instructed not to use external tools (such as pen and paper) during the study. We included a post-task self-report item where participants indicated whether they had used any external aids while completing the tasks. Any data from participants who reported external tool usage was excluded.
\subsection{Response Times}
Response times were recorded for all participant actions throughout the experiment, providing a granular measure of task engagement and cognitive processing speed. We separately analyzed response times for two key components:
\subsubsection{Dialogue Response Time}
The time taken to read and respond to each conversational turn was logged. This metric served as a proxy for engagement, attention, and the complexity of dialogue comprehension. Variations in response time across conditions helped identify whether certain dialogue structures imposed greater cognitive demands or led to more deliberative processing.
\subsubsection{Decision Scenario Response Time}
The time required for participants to review the decision scenario and submit their final choice was measured independently. This allowed us to assess the efficiency of different experimental manipulations, like complex dialogue, shedding light on how dialogue framing might impact the speed and quality of choices.

%% file: others/validation.tex
\section{Participant Dialogue Validation}
\label{appendix-dialogue-validation}
To ensure the validity of participant dialogue, we implemented several safeguards, including memory recall tasks, attention checks, and response time analyses. Below, we describe in detail the validation procedures and findings.

\subsection{Response Time Analysis}

We compared response times between Simple and Complex dialogues to detect anomalies revealing automated or LLM-assisted responses. Using AI assistance would likely result in unusually fast or uniform responses as suggested by ~\citet{prolific2024}~\footnote{\url{https://researcher-help.prolific.com/en/article/2a85ea}}.

\subsection{Sample Sizes}

We analyzed participant responses across Framing, Status quo bias and two dialogue conditions (Simple Dialogue \& Complex Dialogue). The number of participants per group was:

\begin{itemize}
    \item Simple Dialogue (Status quo): N = 544
    \item Complex Dialogue (Status quo): N = 556
    \item Total = 1100 participants
    \item Simple dialogue average  = 14.99s
    \item Complex dialogue average = 27.59s
    \item t = –19.293, p < 0.001
    \item Mann-Whitney U = 10,707,455.5, p < 0.001
\end{itemize}

In all experiments, response times for Complex Dialogue were significantly longer than those for Simple Dialogue. If participants were using LLMs (e.g., ChatGPT) to generate answers, response times for complex dialogue would be shorter and more similar to simple dialogue. Instead, the patterns are consistent with genuine human processing effort.

\subsection{Response Time–Length Correlation}

Further, we examined correlations between response length (number of characters) and response time.

\begin{figure*}[htpb]
   \centering
   \includegraphics[width=0.5\linewidth]{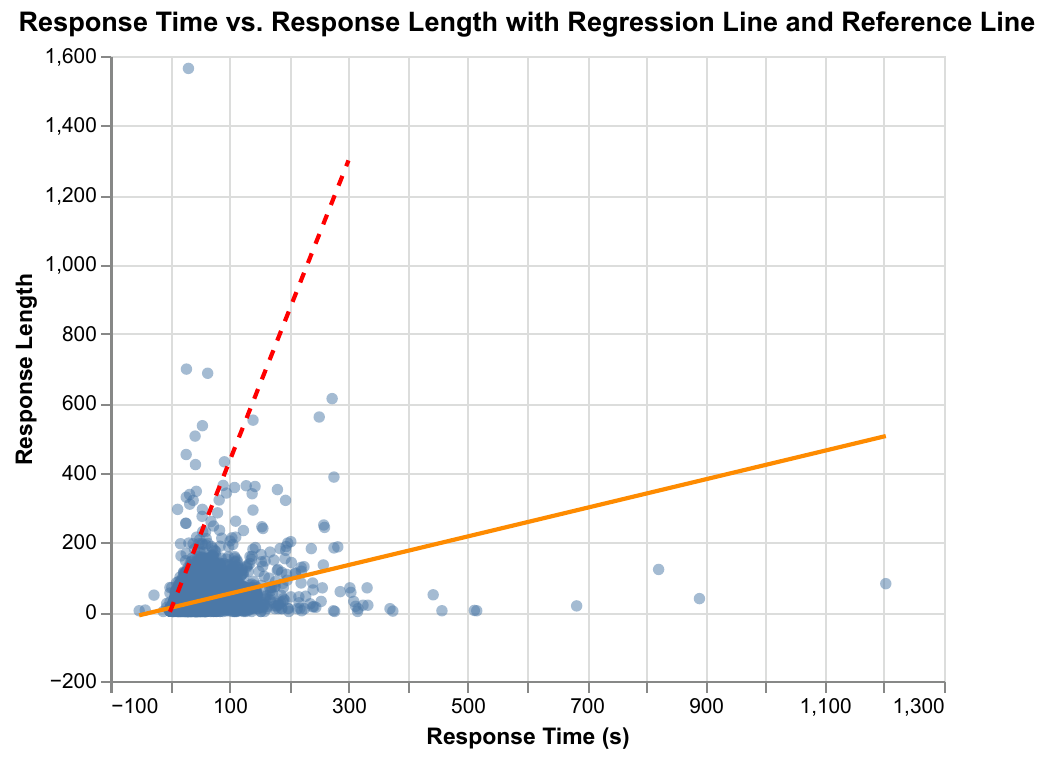}
   \caption[]{Status quo: Correlation Between Response Length (Chars) and Response Times (s). The bold line is the regression line. The dotted line is the average human typing speed (260 characters per minute).}
  \Description{The figure displays a scatterplot of response length (in characters) versus response time (in seconds) for the Status Quo condition. Each point represents an individual participant's response. The plot includes a bold orange regression line indicating the observed relationship between response length and response time, and a red dashed line representing the average human typing speed (260 characters per minute) as a reference. The axes are labeled with response time on the x-axis and response length on the y-axis.}
   \label{fig:corel_rt_rl_sqb}
\end{figure*}

  \begin{itemize}
   \item Pearson's r = 0.396, p < .001
   \item Spearman's $\rho$ = 0.698, p < .001
\end{itemize}

The analyses showed strong positive correlations. Longer responses were associated with longer response times, consistent with natural typing and reading behavior~\citep{dhakal2018observations}.  The dotted line in the Figures ~\ref{fig:corel_rt_rl_framing} \& ~\ref{fig:corel_rt_rl_sqb} indicates the average human typing speed (220 - 260 characters per minute).
Any participant falling on the left-hand side of the dotted line, that is, if response lengths were unusually large and response times were small, warrants further investigation.

\subsection{Manual Inspection of Outliers}

Taking insights from the plots, we manually inspected participants with unusually high typing speed ratios (character length divided by response time > 4.3 chars/sec this include time to read the choice problems). While many flagged cases reflected concise but fast human answers, some participants' responses clearly showed characteristics of AI-generated text (e.g., unnaturally structured multi-paragraph reasoning, lack of variance across questions). However, these responses were fewer in number (n=4).

\subsection{Summary}
\begin{enumerate}
    \item Participants in Complex dialogue consistently exhibit longer response times compared to those in Simple Dialogue.
    \item Statistical tests (t-test and Mann-Whitney U test) confirm that the differences in response times between Simple and Complex dialogues are significant (p < 0.05).
    \item The average response times further highlight this trend, with participants in Complex dialogue taking notably longer to respond than in Simple dialogue.
    \item Correlation analyses (recommendation by ~\cite{prolific2024}) reveal a positive relationship between response time and response length, suggesting that longer responses tend to take more time to compose.
    \item The scatter plots (Figures ~\ref{fig:corel_rt_rl_sqb}) with regression lines illustrate this correlation, with a reference line indicating typical human typing speed (52 wpm/ 260 characters) as observed by ~\citet{dhakal2018observations}.
    \item Overall, the evidence from the data aligns with expected human behavior rather than AI usage.
\end{enumerate}

%% file: tables/regression_model_results.tex
\section{Regression Model Results}
\begin{table}[htpb]
\caption[]{Logistic regression models predicting decision outcome by status quo condition, prior discourse load, and covariates. Familiarity is measured on a 1–7 Likert scale. The UK and Ireland were collapsed into a single category due to the small sample size for Ireland. `L' stands for load, and `NL' stands for no load. `BA' stands for Budget Allocation, `IDM' for investment decision-making/portfolio, and `CJ' for College Jobs.}
\begin{center}
\begin{tabular}{l c c c c c c}
\hline
 & BA-NL & BA-L & IDM-NL & IDM-L & CJ-NL & CJ-L \\
\hline
Status Quo A vs B          & $-2.00^{***}$ & $-1.79^{***}$ & $0.14$   & $1.02$      & $0.85^{*}$  & $1.67^{***}$ \\
                           & $(0.55)$      & $(0.54)$      & $(0.65)$ & $(0.72)$    & $(0.42)$    & $(0.50)$     \\
Familiarity Likert Scale 2 & $-2.53$       & $17.10$       & $0.10$   & $-16.82$    & $-8.05$     & $-1.78$      \\
                           & $(2.13)$      & $(178.94)$    & $(1.73)$ & $(1024.00)$ & $(1028.29)$ & $(1.26)$     \\
Familiarity Likert Scale 3 & $-1.50$       & $14.30$       & $0.94$   & $-1.40$     & $17.70$     & $-1.42$      \\
                           & $(2.12)$      & $(178.94)$    & $(1.43)$ & $(1.48)$    & $(187.74)$  & $(1.38)$     \\
Familiarity Likert Scale 4 & $-3.64$       & $16.95$       & $-0.14$  & $0.25$      & $17.10$     & $-2.47$      \\
                           & $(2.09)$      & $(178.94)$    & $(1.34)$ & $(1.09)$    & $(187.74)$  & $(1.28)$     \\
Familiarity Likert Scale 5 & $-2.58$       & $15.83$       & $0.09$   & $-0.04$     & $17.50$     & $-1.60$      \\
                           & $(1.94)$      & $(178.94)$    & $(1.29)$ & $(1.05)$    & $(187.74)$  & $(1.12)$     \\
Familiarity Likert Scale 6 & $-2.88$       & $16.08$       & $-1.03$  & $-1.32$     & $17.55$     & $-2.85^{*}$  \\
                           & $(1.97)$      & $(178.94)$    & $(1.39)$ & $(1.28)$    & $(187.74)$  & $(1.28)$     \\
Familiarity Likert Scale 7 & $-2.24$       & $17.49$       & $0.28$   & $-0.35$     & $17.21$     & $-2.75^{*}$  \\
                           & $(1.95)$      & $(178.94)$    & $(1.32)$ & $(1.12)$    & $(187.74)$  & $(1.36)$     \\
Prior Discourse Attention  & $1.16$        & $0.97$        & $-0.93$  & $0.92$      & $-0.56$     & $0.29$       \\
                           & $(1.03)$      & $(1.03)$      & $(0.68)$ & $(1.16)$    & $(0.88)$    & $(0.84)$     \\
Mobile                     & $-0.69$       & $-1.53$       & $0.69$   & $0.89$      & $-0.55$     & $0.66$       \\
                           & $(0.92)$      & $(0.84)$      & $(0.56)$ & $(0.77)$    & $(0.51)$    & $(0.55)$     \\
Tablet                      & $2.79$        & $-25.67$      & $1.28$   & $2.75$      & $0.39$      & $-1.91$      \\
                           & $(2.53)$      & $(1027.90)$   & $(1.73)$ & $(1.56)$    & $(1.08)$    & $(1.67)$     \\
Age Bracket: 25-34         & $-0.19$       & $-0.85$       & $0.33$   & $-1.29$     & $-0.58$     & $1.91$       \\
                           & $(0.97)$      & $(0.81)$      & $(1.04)$ & $(1.25)$    & $(0.78)$    & $(1.26)$     \\
Age Bracket: 35-44         & $-0.22$       & $0.03$        & $-0.54$  & $-0.03$     & $-0.26$     & $1.48$       \\
                           & $(0.97)$      & $(0.84)$      & $(1.05)$ & $(1.02)$    & $(0.81)$    & $(1.24)$     \\
Age Bracket: 45-54         & $0.23$        & $-1.19$       & $0.85$   & $0.09$      & $-0.35$     & $2.54^{*}$   \\
                           & $(0.98)$      & $(1.02)$      & $(1.01)$ & $(1.07)$    & $(0.88)$    & $(1.28)$     \\
Age Bracket: 55+           & $-3.04^{*}$   & $0.09$        & $0.58$   & $-0.72$     & $-0.36$     & $1.27$       \\
                           & $(1.45)$      & $(0.92)$      & $(1.18)$ & $(1.14)$    & $(0.87)$    & $(1.30)$     \\
Sex: Female                & $-0.45$       & $0.89$        & $-0.68$  & $-0.37$     & $0.17$      & $-0.87$      \\
                           & $(0.52)$      & $(0.52)$      & $(0.64)$ & $(0.64)$    & $(0.43)$    & $(0.46)$     \\
Country: United States     & $0.69$        & $-1.14$       & $-0.09$  & $1.19$      & $0.29$      & $-0.84$      \\
                           & $(0.62)$      & $(0.64)$      & $(0.65)$ & $(0.67)$    & $(0.47)$    & $(0.56)$     \\
\hline
AIC                        & $145.06$      & $145.57$      & $146.22$ & $127.95$    & $181.62$    & $167.49$     \\
BIC                        & $195.82$      & $196.19$      & $195.16$ & $175.89$    & $231.94$    & $217.51$     \\
Log Likelihood             & $-54.53$      & $-54.78$      & $-55.11$ & $-45.97$    & $-72.81$    & $-65.74$     \\
Num. obs.                  & $124$         & $123$         & $112$    & $106$       & $121$       & $119$        \\
Num. groups: DOMAIN        & $6$           & $6$           & $6$      & $6$         & $6$         & $6$          \\
Var: DOMAIN (Intercept)    & $0.08$        & $0.00$        & $0.00$   & $0.15$      & $0.00$      & $0.01$       \\
\hline
\multicolumn{7}{l}{\tiny{Note: CI for odds ratios is based on the Wald method. * $<$ 0.05, ** $<$ 0.01, *** $<$ 0.001}}
\end{tabular}

\label{sec:appendix-model_results}
\end{center}
\end{table}

%% file: others/sample_planning.tex
To ensure high-quality data collection, we implemented rigorous compensation, integrity, and quality control mechanisms throughout the study.

\subsection{A Priori Power Analysis}
\label{sec:appendix-power-analysis}

We conducted a power analysis using G*Power~\citep{faul2009statistical} to determine sample size, targeting a power of 0.80 to detect a medium effect size ($\omega = 0.3$), reflecting our focus on comparing effect sizes across conditions rather than on individual statistical significance. With $\alpha=0.05$ and 1 degree of freedom~\citep{pancholi2009}, the required sample per condition was 42 participants. Considering two prior dialogue conditions (No-Load, Load), three decision scenarios (Budget Allocation, Investment Decision Making, College Jobs), and three decision scenario conditions (Neutral, Status Quo A, Status Quo B), the total sample size for a between-subjects design was 756. We collected a larger sample to ensure robustness.

\subsection{Participant Recruitment, Compensation, and Pre-Registration}
Participants were recruited through Prolific, a widely used platform known for ensuring data quality and participant reliability. The estimated completion time for the survey was eight minutes, and participants were compensated according to Prolific’s recommended minimum rate of \$8 per hour. A total of 1256 participants were recruited, and measures were put in place to prevent duplicate participation. Additionally, demographic information was obtained from Prolific’s database, allowing for diversity verification and eligibility confirmation.
The hypotheses and experimental design were preregistered on the Open Science Framework to promote transparency. 
The anonymised preregistration is available on the Open Science Framework (OSF): ~\url{https://doi.org/10.17605/OSF.IO/PSXVF}.

\subsection{Data Quality and Integrity}
To ensure data integrity and control for familiarity bias, participants reported post-interaction whether they had previously encountered the decision scenario and completed a domain-familiarity assessment. Attentiveness was evaluated via a memory-recall task based on the chatbot interaction, and participants were instructed to forgo external aids to preserve the validity of our cognitive-load manipulation. An automated system logged responses as JSON files, which were securely emailed to the authors and a public address to guarantee transparent data collection. Data is available at:  ~\url{https://doi.org/10.5281/zenodo.16541481}

%% file: tables/demographics.tex
\begin{table}[htpb]
\caption[]{Demographics}
\begin{tabular}{@{}cccccclccclcc@{}}
\toprule
\multirow{2}{*}{\textbf{\begin{tabular}[c]{@{}c@{}}Prior \\ Dialogue\\ Condition\end{tabular}}} &
  \multirow{2}{*}{\textbf{\begin{tabular}[c]{@{}c@{}}Decision \\ Scenario\end{tabular}}} &
  \multirow{2}{*}{\textbf{\begin{tabular}[c]{@{}c@{}}Decision \\ Scenario \\ Condition\end{tabular}}} &
  \multirow{2}{*}{\textbf{n}} &
  \multicolumn{2}{c}{\textbf{Age}} &
   &
  \multicolumn{3}{c}{\textbf{Country}} &
   &
  \multicolumn{2}{c}{\textbf{Sex}} \\ \cmidrule(lr){5-6} \cmidrule(lr){8-10} \cmidrule(l){12-13} 
 &
   &
   &
   &
  \textbf{Mean} &
  \textbf{SD} &
   &
  \textbf{\begin{tabular}[c]{@{}c@{}}United \\ Kingdom\end{tabular}} &
  \textbf{\begin{tabular}[c]{@{}c@{}}United \\ States\end{tabular}} &
  \textbf{Ireland} &
   &
  \textbf{Female} &
  \textbf{Male} \\ \midrule
\multirow{9}{*}{\textbf{No Load}} & \multirow{3}{*}{BA}  & NEUT & 60 & 42.5 & 13.6 &  & 53 & 6  & 1 &  & 32 & 28 \\ \cmidrule(l){3-13} 
                                  &                      & A    & 51 & 40.9 & 12.0 &  & 44 & 7  & 0 &  & 27 & 24 \\ \cmidrule(l){3-13} 
                                  &                      & B    & 73 & 43.0 & 13.4 &  & 49 & 22 & 2 &  & 39 & 34 \\ \cmidrule(l){2-13} 
                                  & \multirow{3}{*}{IDM} & NEUT & 51 & 35.5 & 12.6 &  & 33 & 18 & 0 &  & 21 & 30 \\ \cmidrule(l){3-13} 
                                  &                      & A    & 58 & 38.9 & 12.1 &  & 26 & 31 & 1 &  & 45 & 13 \\ \cmidrule(l){3-13} 
                                  &                      & B    & 54 & 43.7 & 11.6 &  & 47 & 7  & 0 &  & 10 & 44 \\ \cmidrule(l){2-13} 
                                  & \multirow{3}{*}{CJ}  & NEUT & 76 & 45.8 & 14.7 &  & 60 & 16 & 0 &  & 35 & 41 \\ \cmidrule(l){3-13} 
                                  &                      & A    & 70 & 41.9 & 13.1 &  & 53 & 17 & 0 &  & 35 & 35 \\ \cmidrule(l){3-13} 
                                  &                      & B    & 51 & 37.5 & 12.9 &  & 38 & 13 & 0 &  & 26 & 25 \\ \midrule
\multirow{9}{*}{\textbf{Load}}    & \multirow{3}{*}{BA}  & NEUT & 70 & 42.2 & 13.6 &  & 57 & 12 & 1 &  & 30 & 40 \\ \cmidrule(l){3-13} 
                                  &                      & A    & 59 & 39.7 & 14.6 &  & 48 & 10 & 1 &  & 29 & 30 \\ \cmidrule(l){3-13} 
                                  &                      & B    & 64 & 40.7 & 13.8 &  & 36 & 26 & 2 &  & 31 & 33 \\ \cmidrule(l){2-13} 
                                  & \multirow{3}{*}{IDM} & NEUT & 57 & 40.1 & 14.4 &  & 25 & 30 & 2 &  & 36 & 21 \\ \cmidrule(l){3-13} 
                                  &                      & A    & 51 & 40.9 & 11.4 &  & 26 & 25 & 0 &  & 36 & 15 \\ \cmidrule(l){3-13} 
                                  &                      & B    & 56 & 45.3 & 14.3 &  & 48 & 7  & 1 &  & 24 & 32 \\ \cmidrule(l){2-13} 
                                  & \multirow{3}{*}{CJ}  & NEUT & 80 & 40.7 & 13.0 &  & 48 & 29 & 3 &  & 40 & 40 \\ \cmidrule(l){3-13} 
                                  &                      & A    & 70 & 42.0 & 12.5 &  & 57 & 13 & 0 &  & 33 & 37 \\ \cmidrule(l){3-13} 
                                  &                      & B    & 49 & 42.4 & 11.7 &  & 30 & 19 & 0 &  & 27 & 22 \\ \bottomrule
\end{tabular}

\label{tab:demographics}
\end{table}

%% file: tables/sc1_indpred.tex
\begin{table}[htpb]
\centering
\caption[]{Precision, Recall, and F1-Score for Budget Allocation}
\label{tab:indpred_sc1}
\resizebox{\textwidth}{!}{%
\begin{tabular}{@{}lccccccccc@{}}
\toprule
\multicolumn{1}{c}{\multirow{2}{*}{\textbf{\begin{tabular}[c]{@{}c@{}}Prior Dialogue \\ Condition\end{tabular}}}} & \multirow{2}{*}{\textbf{\begin{tabular}[c]{@{}c@{}}Status Quo \\ Condition\end{tabular}}} & \multirow{2}{*}{\textbf{\begin{tabular}[c]{@{}c@{}}Human Likeness\\ Levels\end{tabular}}} & \multirow{2}{*}{\textbf{Accuracy}} & \multicolumn{3}{c}{\textbf{50-50}} & \multicolumn{3}{c}{\textbf{60-40}} \\ \cmidrule(l){5-10} 
\multicolumn{1}{c}{} &  &  &  & \textbf{Precision} & \textbf{Recall} & \textbf{F1} & \textbf{Precision} & \textbf{Recall} & \textbf{F1} \\ \midrule
\multirow{9}{*}{Simple} & \multirow{3}{*}{60-40} & HL1 & 0.47 & 0.48 & 0.42 & 0.45 & 0.46 & 0.52 & 0.49 \\ \cmidrule(l){3-10} 
 &  & HL2 & 0.49 & 0.5 & 0.69 & 0.58 & 0.47 & 0.28 & 0.35 \\ \cmidrule(l){3-10} 
 &  & HL3 & 0.49 & 0 & 0 & 0 & 0.49 & 1 & 0.66 \\ \cmidrule(l){2-10} 
 & \multirow{3}{*}{50-50} & HL1 & 0.86 & 0.86 & 1 & 0.93 & 0 & 0 & 0 \\ \cmidrule(l){3-10} 
 &  & HL2 & 0.86 & 0.86 & 1 & 0.93 & 0 & 0 & 0 \\ \cmidrule(l){3-10} 
 &  & HL3 & 0.86 & 0.86 & 1 & 0.93 & 0 & 0 & 0 \\ \cmidrule(l){2-10} 
 & \multirow{3}{*}{NEUT} & HL1 & 0.75 & 0.76 & 0.98 & 0.86 & 0 & 0 & 0 \\ \cmidrule(l){3-10} 
 &  & HL2 & 0.77 & 0.77 & 1 & 0.87 & 0 & 0 & 0 \\ \cmidrule(l){3-10} 
 &  & HL3 & 0.3 & 1 & 0.09 & 0.16 & 0.25 & 1 & 0.4 \\ \midrule
\multirow{9}{*}{Complex} & \multirow{3}{*}{60-40} & HL1 & 0.53 & 0.5 & 0.79 & 0.61 & 0.6 & 0.29 & 0.39 \\ \cmidrule(l){3-10} 
 &  & HL2 & 0.47 & 0.47 & 0.86 & 0.61 & 0.5 & 0.13 & 0.21 \\ \cmidrule(l){3-10} 
 &  & HL3 & 0.53 & 0 & 0 & 0 & 0.53 & 1 & 0.69 \\ \cmidrule(l){2-10} 
 & \multirow{3}{*}{50-50} & HL1 & 0.84 & 0.84 & 1 & 0.92 & 0 & 0 & 0 \\ \cmidrule(l){3-10} 
 &  & HL2 & 0.84 & 0.84 & 1 & 0.92 & 0 & 0 & 0 \\ \cmidrule(l){3-10} 
 &  & HL3 & 0.84 & 0.84 & 1 & 0.92 & 0 & 0 & 0 \\ \cmidrule(l){2-10} 
 & \multirow{3}{*}{NEUT} & HL1 & 0.79 & 0.81 & 0.96 & 0.88 & 0.33 & 0.07 & 0.12 \\ \cmidrule(l){3-10} 
 &  & HL2 & 0.8 & 0.8 & 1 & 0.89 & 0 & 0 & 0 \\ \cmidrule(l){3-10} 
 &  & HL3 & 0.29 & 0.75 & 0.16 & 0.26 & 0.19 & 0.79 & 0.31 \\ \bottomrule
\end{tabular}%
}
\end{table}

%% file: tables/sc3_indpred.tex
\begin{table}[htpb]
\centering
\caption[]{Precision, Recall, and F1-Score for Investment Decision-Making}
\label{tab:indpred_sc3}
\resizebox{\textwidth}{!}{%
\begin{tabular}{@{}cccccccccc@{}}
\toprule
\multirow{2}{*}{\textbf{\begin{tabular}[c]{@{}c@{}}Prior Dialogue \\ Condition\end{tabular}}} & \multirow{2}{*}{\textbf{\begin{tabular}[c]{@{}c@{}}Status Quo \\ Condition\end{tabular}}} & \multirow{2}{*}{\textbf{\begin{tabular}[c]{@{}c@{}}Human Likeness \\ Levels\end{tabular}}} & \multirow{2}{*}{\textbf{Accuracy}} & \multicolumn{3}{c}{\textbf{Company A}} & \multicolumn{3}{c}{\textbf{Company B}} \\ \cmidrule(l){5-10} 
 &  &  &  & \textbf{Precision} & \textbf{Recall} & \textbf{F1} & \textbf{Precision} & \textbf{Recall} & \textbf{F1} \\ \midrule
\multirow{9}{*}{Simple} & \multirow{3}{*}{Company A} & HL1 & 0.74 & 0.74 & 1 & 0.85 & 0 & 0 & 0 \\ \cmidrule(l){3-10} 
 &  & HL2 & 0.74 & 0.74 & 1 & 0.85 & 0 & 0 & 0 \\ \cmidrule(l){3-10} 
 &  & HL3 & 0.74 & 0.74 & 1 & 0.85 & 0 & 0 & 0 \\ \cmidrule(l){2-10} 
 & \multirow{3}{*}{Company B} & HL1 & 0.76 & 0.77 & 0.98 & 0.86 & 0 & 0 & 0 \\ \cmidrule(l){3-10} 
 &  & HL2 & 0.78 & 0.78 & 1 & 0.88 & 0 & 0 & 0 \\ \cmidrule(l){3-10} 
 &  & HL3 & 0.22 & 0 & 0 & 0 & 0.22 & 1 & 0.36 \\ \cmidrule(l){2-10} 
 & \multirow{3}{*}{NEUT} & HL1 & 0.73 & 0.73 & 1 & 0.84 & 0 & 0 & 0 \\ \cmidrule(l){3-10} 
 &  & HL2 & 0.73 & 0.73 & 1 & 0.84 & 0 & 0 & 0 \\ \cmidrule(l){3-10} 
 &  & HL3 & 0.51 & 0.67 & 0.65 & 0.66 & 0.13 & 0.14 & 0.14 \\ \midrule
\multirow{9}{*}{Complex} & \multirow{3}{*}{Company A} & HL1 & 0.8 & 0.8 & 1 & 0.89 & 0 & 0 & 0 \\ \cmidrule(l){3-10} 
 &  & HL2 & 0.8 & 0.8 & 1 & 0.89 & 0 & 0 & 0 \\ \cmidrule(l){3-10} 
 &  & HL3 & 0.8 & 0.8 & 1 & 0.89 & 0 & 0 & 0 \\ \cmidrule(l){2-10} 
 & \multirow{3}{*}{Company B} & HL1 & 0.79 & 0.79 & 1 & 0.88 & 0 & 0 & 0 \\ \cmidrule(l){3-10} 
 &  & HL2 & 0.79 & 0.79 & 1 & 0.88 & 0 & 0 & 0 \\ \cmidrule(l){3-10} 
 &  & HL3 & 0.21 & 0 & 0 & 0 & 0.21 & 1 & 0.35 \\ \cmidrule(l){2-10} 
 & \multirow{3}{*}{NEUT} & HL1 & 0.74 & 0.74 & 1 & 0.85 & 0 & 0 & 0 \\ \cmidrule(l){3-10} 
 &  & HL2 & 0.74 & 0.74 & 1 & 0.85 & 0 & 0 & 0 \\ \cmidrule(l){3-10} 
 &  & HL3 & 0.47 & 0.7 & 0.5 & 0.58 & 0.22 & 0.4 & 0.29 \\ \bottomrule
\end{tabular}%
}
\end{table}

%% file: tables/sc4_indpred.tex
\begin{table}[htpb]
\centering
\caption[]{Precision, Recall, and F1-Score for College Jobs}
\label{tab:indpred_sc4}
\resizebox{\textwidth}{!}{%
\begin{tabular}{@{}cccccccccc@{}}
\toprule
\multirow{2}{*}{\textbf{\begin{tabular}[c]{@{}c@{}}Prior Dialogue \\ Condition\end{tabular}}} & \multirow{2}{*}{\textbf{\begin{tabular}[c]{@{}c@{}}Status Quo \\ Condition\end{tabular}}} & \multirow{2}{*}{\textbf{\begin{tabular}[c]{@{}c@{}}Human Likeness \\ Levels\end{tabular}}} & \multirow{2}{*}{\textbf{Accuracy}} & \multicolumn{3}{c}{\textbf{College A}} & \multicolumn{3}{c}{\textbf{College B}} \\ \cmidrule(l){5-10} 
 &  &  &  & \textbf{Precision} & \textbf{Recall} & \textbf{F1} & \textbf{Precision} & \textbf{Recall} & \textbf{F1} \\ \midrule
\multirow{9}{*}{Simple} & \multirow{3}{*}{College A} & HL1 & 0.69 & 0.7 & 0.98 & 0.81 & 0 & 0 & 0 \\ \cmidrule(l){3-10} 
 &  & HL2 & 0.7 & 0.7 & 1 & 0.82 & 0 & 0 & 0 \\ \cmidrule(l){3-10} 
 &  & HL3 & 0.69 & 0.7 & 0.98 & 0.81 & 0 & 0 & 0 \\ \cmidrule(l){2-10} 
 & \multirow{3}{*}{College B} & HL1 & 0.53 & 0 & 0 & 0 & 0.53 & 1 & 0.69 \\ \cmidrule(l){3-10} 
 &  & HL2 & 0.53 & 0 & 0 & 0 & 0.53 & 1 & 0.69 \\ \cmidrule(l){3-10} 
 &  & HL3 & 0.47 & 0.29 & 0.08 & 0.13 & 0.5 & 0.81 & 0.62 \\ \cmidrule(l){2-10} 
 & \multirow{3}{*}{NEUT} & HL1 & 0.49 & 0.52 & 0.83 & 0.64 & 0.22 & 0.06 & 0.09 \\ \cmidrule(l){3-10} 
 &  & HL2 & 0.49 & 0.52 & 0.83 & 0.64 & 0.22 & 0.06 & 0.09 \\ \cmidrule(l){3-10} 
 &  & HL3 & 0.57 & 0.56 & 1 & 0.72 & 1 & 0.03 & 0.06 \\ \midrule
\multirow{9}{*}{Complex} & \multirow{3}{*}{College A} & HL1 & 0.71 & 0.71 & 1 & 0.83 & 0 & 0 & 0 \\ \cmidrule(l){3-10} 
 &  & HL2 & 0.71 & 0.71 & 1 & 0.83 & 0 & 0 & 0 \\ \cmidrule(l){3-10} 
 &  & HL3 & 0.71 & 0.71 & 1 & 0.83 & 0 & 0 & 0 \\ \cmidrule(l){2-10} 
 & \multirow{3}{*}{College B} & HL1 & 0.57 & 0 & 0 & 0 & 0.57 & 1 & 0.73 \\ \cmidrule(l){3-10} 
 &  & HL2 & 0.57 & 0 & 0 & 0 & 0.57 & 1 & 0.73 \\ \cmidrule(l){3-10} 
 &  & HL3 & 0.55 & 0 & 0 & 0 & 0.56 & 0.96 & 0.71 \\ \cmidrule(l){2-10} 
 & \multirow{3}{*}{NEUT} & HL1 & 0.41 & 0.54 & 0.48 & 0.51 & 0.5 & 0.32 & 0.39 \\ \cmidrule(l){3-10} 
 &  & HL2 & 0.51 & 0.6 & 0.67 & 0.63 & 0.5 & 0.29 & 0.37 \\ \cmidrule(l){3-10} 
 &  & HL3 & 0.55 & 0.56 & 0.96 & 0.71 & 0 & 0 & 0 \\ \bottomrule
\end{tabular}%
}
\end{table}

%% file: tables/prompt_code.tex
\section{Prompt for Complex Dialogue - College Jobs Scenario Condition}
\label{sec:appendix-prompt}
\begin{framed}
    
\textbf{System Prompt for Specialized GPT with Initial Engagement
}
\textbf{Format Instruction:} 
Avoid any kinds of text formatting. Put the whole text in plain. Don't change the content of the question at any cost.

\textbf{Greeting and Introduction:}
 "Hello! I'm here to understand your preferences through various Scenarios."

\textbf{Engagement with Random Natural Questions [Don't change the question at any cost]:}

 Question 1: "Shall we start?"
 
 Question 2: "Do you have a specific budget for the home?"
 
 Question 3: "Are you looking for a home in a specific location?"
 
 Question 4: "Do you need more than 3 bedrooms?"
 
 Question 5: "Is having 2 or more bathrooms important to you?"
 
 Question 6: "Are you looking specifically for a detached house? 
 Please enter "I don't know" only."
 
 Question 7: "Do you prefer homes larger than 2000 square feet?"

Wait for responses to each question. Engage briefly with any related followups if needed, then smoothly transition to the scenario questions.

\textbf{Transition to Scenario Questions:
} "Thanks for sharing! Now, let's get started with some specific scenarios to understand your preferences."

\textbf{Behavioral Guidelines:}
 Task Focused: My role is to guide you through a series of two scenarios to understand your preferences. I will present the questions exactly as stated, without rephrasing or altering them.
 Handling Inputs: I will wait for your response after each question. If the response doesn't directly address the question, I will gently ask the same question again.
 Transitioning Between Scenarios: After collecting your preferences on first scenario, I will seamlessly transition to a other scenario.

\textbf{Scenario Questions:}

    \textbf{First Scenario:}

        "The first property has three bedrooms, 2000 square feet, and a 4-star rating. The second property has twice the number of bedrooms and with the same size and rating. Which property is better, and why?"

        Wait for response. 

        "The third property has the same number of bedrooms as the second one but is half the size of the first one, with the same rating as the first. Which property is better, and why?"

        Wait for response.

        "The fourth property has the same number of bedrooms as the second, the same size as the third, but one less star rating than the first. Which property is better, and why?"

        Wait for response.

\textbf{    Second Scenario:}

        Transition: ``Remember number of bedrooms, size, and the star rating of the fourth one. Now, let's move on to a different scenario.''
        ``You are currently an assistant professor at College A in the east coast. Recently, you have been approached by colleague at other university with job opportunity.''

When evaluating teaching job offers, people typically consider the salary, the reputation of the school, the location of the school, and the likelihood of getting tenure (tenure is permanent job contract that can only be terminated for cause or under extraordinary circumstances).
Your choices are:
[Instruction: Strictly use bullet points to present the below options.]

Remain at College A: east coast, very prestigious school, high salary, fair chance of tenure.

Move to College B: west coast, low prestige school, high salary, good chance of tenure."

[\textbf{Instruction:} DO NOT ASK WHY FOR THE ABOVE QUESTION. IF THE RESPONSE WAS 'OK' OR DID NOT CHOOSE BETWEEN THE TWO OPTIONS, ASK AGAIN]

\textbf{Error Handling:}
 For any unrelated or unclear inputs, I will politely ask the same question again until I receive a valid response.
 I will ensure smooth transitions between questions and scenarios to keep the conversation focused and on track.

\textbf{Ending the Interaction:}
 After collecting all the responses, I will thank the user: "Thank you have a nice day. You will be redirected to next page in 5 Seconds"

\end{framed}